\documentclass{ieee-ojvt}
\usepackage{amsmath,amssymb,amsfonts}
\usepackage{algorithmic}
\usepackage{graphicx}
\usepackage{textcomp}
\usepackage{hyperref} 
\usepackage[numbers,sort&compress]{natbib}

\usepackage{xcolor}

\usepackage{tabularx, multirow, booktabs, makecell, array}
\newcolumntype{M}[1]{>{\centering\arraybackslash}m{#1}}
\newcolumntype{L}[1]{>{\raggedright\arraybackslash}p{#1}}
\def\BibTeX{{\rm B\kern-.05em{\sc i\kern-.025em b}\kern-.08em
    T\kern-.1667em\lower.7ex\hbox{E}\kern-.125emX}}
\AtBeginDocument{\definecolor{ojcolor}{cmyk}{0.93,0.59,0.15,0.02}}
\def\OJlogo{\vspace{-12pt}\includegraphics[height=24pt]{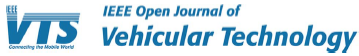}}

\begin{document}
\receiveddate{XX Month, XXXX}
\reviseddate{XX Month, XXXX}
\accepteddate{XX Month, XXXX}
\publisheddate{XX Month, XXXX}
\currentdate{27 June, 2024}
\doiinfo{OJVT.2024.0627000}

\title{Foundation Models for Autonomous Driving Perception: A Survey Through Core Capabilities  }

\author{%
  RAJENDRAMAYAVAN SATHYAM\authorrefmark{1}\textsuperscript{1}, 
  YUEQI LI\authorrefmark{1}\textsuperscript{1} 
}
\affil{Zoox Inc.}

\markboth{Preparation of Papers for IEEE OPEN JOURNALS}{Author \textit{et al.}}

\begin{abstract}
Foundation models are revolutionizing autonomous driving perception, transitioning the field from narrow, task-specific deep learning models to versatile, general-purpose architectures trained on vast, diverse datasets. This survey examines how these models address critical challenges in autonomous perception, including limitations in generalization, scalability, and robustness to distributional shifts. The survey introduces a novel taxonomy structured around four essential capabilities for robust performance in dynamic driving environments: generalized knowledge, spatial understanding, multi-sensor robustness, and temporal reasoning. For each capability, the survey elucidates its significance and comprehensively reviews cutting-edge approaches. Diverging from traditional method-centric surveys, our unique framework prioritizes conceptual design principles, providing a capability-driven guide for model development and clearer insights into foundational aspects. We conclude by discussing key challenges, particularly those associated with the integration of these capabilities into real-time, scalable systems, and broader deployment challenges related to computational demands and ensuring model reliability against issues like hallucinations and out-of-distribution failures. The survey also outlines crucial future research directions to enable the safe and effective deployment of foundation models in autonomous driving systems.  
\end{abstract}



\maketitle

\begingroup
\renewcommand\thefootnote{}\footnotetext{\textsuperscript{*}Equal contribution}
\addtocounter{footnote}{-1}
\endgroup

\section{\MakeUppercase{INTRODUCTION}}
\IEEEPARstart{A}{utonomous} driving perception constitutes a cornerstone of intelligent transportation systems, as it is responsible for enabling vehicles to interpret and respond to their surrounding environments in real time. The perception stack involves a suite of tasks such as object detection, semantic segmentation, and object tracking—each of which is essential for safe navigation and effective decision-making \cite{qian20223d, muhammad2022vision, guo2022review}. Traditionally, these tasks have been tackled using narrowly focused deep learning models designed explicitly for individual perception tasks, which are trained on meticulously curated, manually annotated datasets. While achieving impressive performance within controlled scenarios, such specialized models inherently suffer from limited scalability and poor generalizability. They frequently experience significant performance degradation when exposed to long-tailed scenarios, which include infrequent yet safety-critical events, or rare occurrences insufficiently represented in training data \cite{peri2023towards}. Moreover, these models often struggle to adapt effectively when faced with distributional shifts arising from changes in environmental conditions, sensor configurations, or operational contexts.

Recent breakthroughs in the development of foundation models have initiated a transformative shift in autonomous driving perception. Foundation models refer to large-scale, general-purpose neural networks that are pre-trained on extensive, diverse datasets using self-supervised or unsupervised learning strategies \cite{bommasani2021opportunities}. Typically, these models are built by leveraging architectures such as transformers, which effectively model complex interactions in data, allowing them to learn universal representations and capture latent knowledge about the world without explicit supervision. The pre-training process involves exposing these models to vast quantities of data from multiple modalities—text, images, videos, and potentially other sensor data—enabling them to acquire generalized features useful across various downstream tasks.

In the context of autonomous driving, foundation models provide significant advantages due to their inherent capability for broad generalization, efficient transfer learning, and reduced reliance on task-specific annotated datasets. By possessing extensive general knowledge encoded during the pre-training phase, these models can more readily adapt to diverse and dynamic real-world driving conditions, improving performance in scenarios beyond the reach of traditional supervised approaches. Moreover, the unified representation learned by foundation models facilitates seamless integration across perception tasks, promoting consistency and coherence in interpreting complex driving environments.

Foundation models need extensive data and large-scale architectures to encode broad and generalizable knowledge. In the context of autonomous driving, it is important not only to scale model size and data volume, but also to apply specialized techniques and deliberate training strategies. These methods help guide the model to acquire capabilities that are particularly relevant to the unique challenges of autonomous vehicle perception. Recognizing this need, this paper identifies and elaborates on four critical aspects that are particularly essential for constructing and refining foundation models tailored specifically for autonomous driving perception:

\begin{enumerate}
    \item \textbf{Generalized knowledge.} The model should be capable of adapting to a broad range of driving scenarios, including rare or previously unseen situations. It must infer plausible outcomes and reason about unfamiliar agents in a principled manner.
    \item \textbf{Spatial awareness.} An effective model must possess a strong understanding of 3D spatial structure and relationships. This includes detecting both known and unknown objects, as well as reasoning about their physical interactions and future trajectories.
    \item \textbf{Multi-sensor robustness.} The system should maintain high performance across a wide spectrum of environmental conditions, such as varying weather, illumination, and sensor noise. Robustness to partial sensor failures is also critical for ensuring safety.
    \item \textbf{Temporal understanding.} Beyond instantaneous perception, the model must capture temporal dependencies and anticipate future states of the environment. This entails modeling motion patterns, recognizing occluded agents, and reasoning about object permanence.
\end{enumerate}

\begin{figure*}[htbp]
    \centering
    \includegraphics[width=\textwidth]{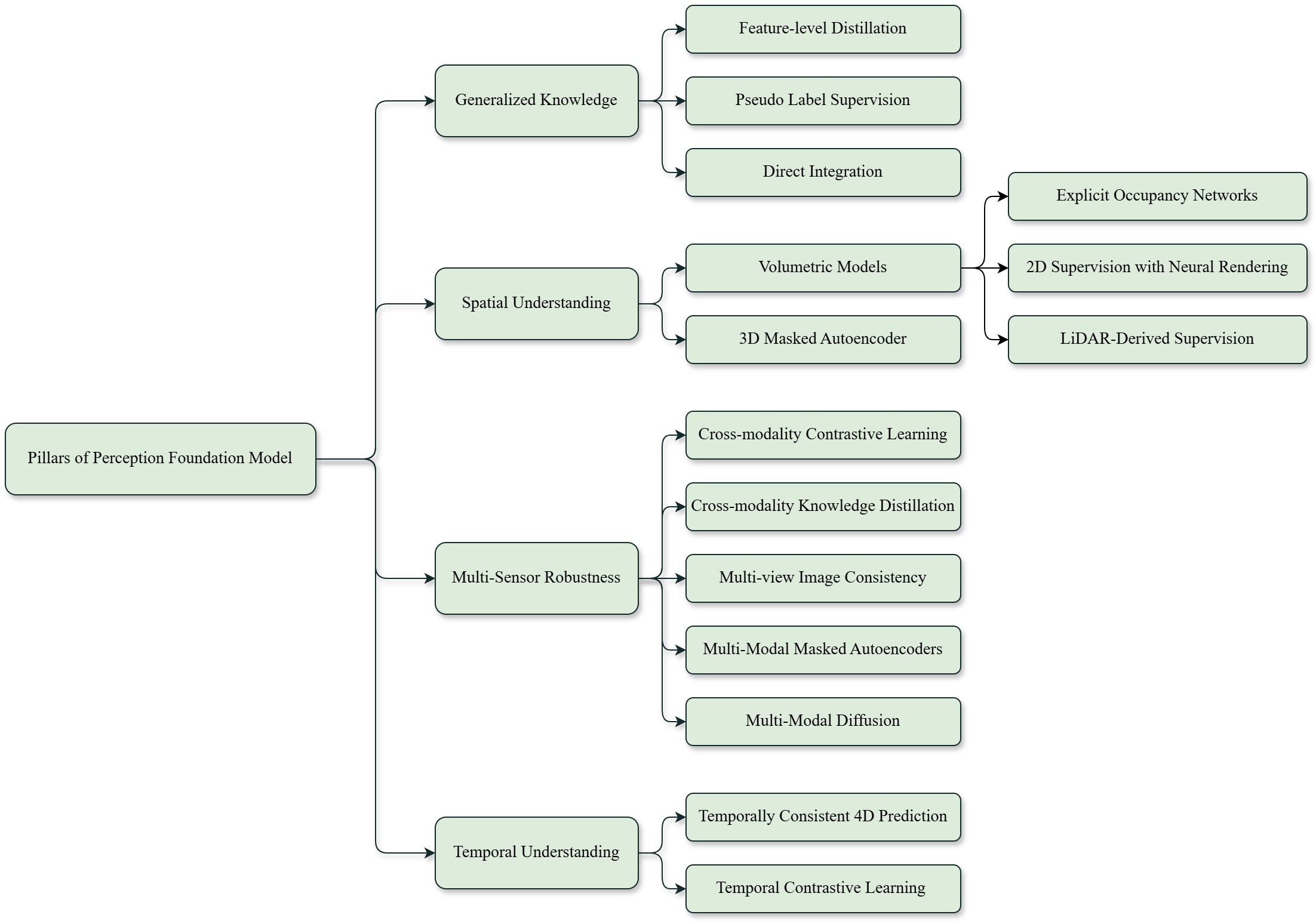}
    \caption{Overview of the four key pillars for foundation models in autonomous driving and the corresponding methods to achieve them.}
    \label{fig:overview}
\end{figure*}

For each of these aspects, we first clearly outline its significance in the context of foundation models for autonomous driving perception. Subsequently, we provide a comprehensive analysis of the underlying key techniques and methodologies employed to effectively develop and enhance the respective capabilities. These techniques are illustrated in \autoref{fig:overview}. We provide abundant examples throughout to clarify their practical implementation and highlight their relevance in real-world autonomous driving scenarios. Finally, we critically analyze current challenges associated with improving these aspects and propose potential directions for future research.

As summarized in Table \ref{survey}, existing surveys on foundation models in autonomous driving predominantly adopt method-based or task-based taxonomies \cite{yan2024forging, huang2023applications, gao2024survey, wu2024prospective, yang2023llm4drive, zhou2024vision}. While valuable, these approaches often focus on a wide array of applications beyond perception, such as prediction and planning. Consequently, they do not provide a comprehensive treatment of all the core capabilities essential for robust perception; critical aspects like multi-sensor robustness and spatial awareness are frequently overlooked. In contrast, our survey introduces a capability-based framework tailored specifically to the perception domain. By structuring the discussion around the four fundamental capabilities we have identified, our taxonomy offers a more focused and in-depth analysis. This approach not only clarifies the essential components of perception-centric foundation models but also provides a systematic guide for researchers to identify and address specific deficiencies, thereby fostering targeted innovation in this critical area.

\begin{table*}[t]
  \centering
  \caption{Survey Papers on Foundation Models for Autonomous Driving}
  \label{survey}
  \begin{tabular}{
      >{\centering\arraybackslash}p{0.25\linewidth}|%
      >{\centering\arraybackslash}p{0.15\linewidth}|%
      >{\centering\arraybackslash}p{0.25\linewidth}|%
      >{\centering\arraybackslash}p{0.25\linewidth}}
    \hline
    \textbf{Survey} & \textbf{Taxonomy}& \textbf{Covered Capabilities} & \textbf{Task(s) in Autonomous driving}\\
    \hline
    Forging Vision Foundation Models for Autonomous Driving  \cite{yan2024forging}& Method‑based & Generalized knowledge, Multi‑sensor robustness, Spatial awareness, Temporal understanding & Perception, simulation \\
    \hline
    Applications of Large Scale Foundation Models for Autonomous Driving \cite{huang2023applications}& Method‑based \& task‑based & Generalized knowledge, Spatial awareness, Temporal understanding & Perception, prediction, planning/control, end‑to‑end, simulation\\
    \hline
    A Survey for Foundation Models in Autonomous Driving \cite{gao2024survey}& Method‑based \& task‑based & Generalized knowledge, Temporal understanding & Perception, prediction, planning/control, end‑to‑end, simulation\\
    \hline
    Prospective Role of Foundation Models in Advancing Autonomous Vehicles \cite{wu2024prospective}& Method‑based \& task‑based & Generalized knowledge, Temporal understanding & Perception, Prediction, planning/control, simulation, end‑to‑end\\
    \hline
    LLM4Drive: A Survey of Large Language Models for Autonomous Driving \cite{yang2023llm4drive}& Task‑based & Generalized knowledge & Perception, planning/control, simulation \\
    \hline
    Vision Language Models in Autonomous Driving: A Survey and Outlook \cite{zhou2024vision}& Task‑based & Generalized knowledge & Perception, planning/control, end‑to‑end \\
    \hline
    Ours & Capability‑based & Generalized knowledge, Multi‑sensor robustness, Spatial awareness, Temporal understanding & Perception \\
    \hline
  \end{tabular}
  \label{tab:fm_surveys_overview}
\end{table*}

\section{\MakeUppercase{Background}}
This section provides essential background on several key topics that underpin the discussion in the paper.  
\subsection{\MakeUppercase{Self Supervised learning}}
Self-supervised learning has emerged as a powerful paradigm for training a foundation model, reducing the reliance on expensive labeled data. Foundation models aim to learn broad, reusable representations that can generalize across tasks and domains, often requiring pretraining on large and diverse datasets \cite{bommasani2021opportunities}. However, acquiring fine-grained annotations at this scale is typically infeasible. Self-supervised learning addresses this challenge by exploiting intrinsic structures or patterns within the data itself as a source of supervision. This allows models to learn high-quality representations from raw input data, fostering scalability and cross-domain adaptability. Consequently, self-supervised methods have become foundational for training models that are intended to serve as a universal backbone for a variety of downstream applications.

Contrastive learning has become a cornerstone of self-supervised representation learning, enabling models to extract semantically meaningful features from unlabeled data by contrasting similar and dissimilar examples \cite{chen2020simple}. The fundamental idea is to train a model to bring representations of similar ("positive") sample pairs closer in the embedding space, while pushing apart dissimilar ("negative") pairs. For example, in the domain of computer vision, contrastive learning techniques such as SimCLR \cite{chen2020simple} and MoCo \cite{he2020momentum,chen2020improved}, learn robust and semantically meaningful representations by maximizing agreement between different augmented views of the same image—such as crops, color distortions, or geometric transformations—while contrasting them against views from different images. This encourages the model to focus on high-level features that are invariant to superficial changes, fostering transferability across tasks such as classification, detection, and segmentation.

\begin{figure}[htbp]
    \centering
    \includegraphics[width=\columnwidth]{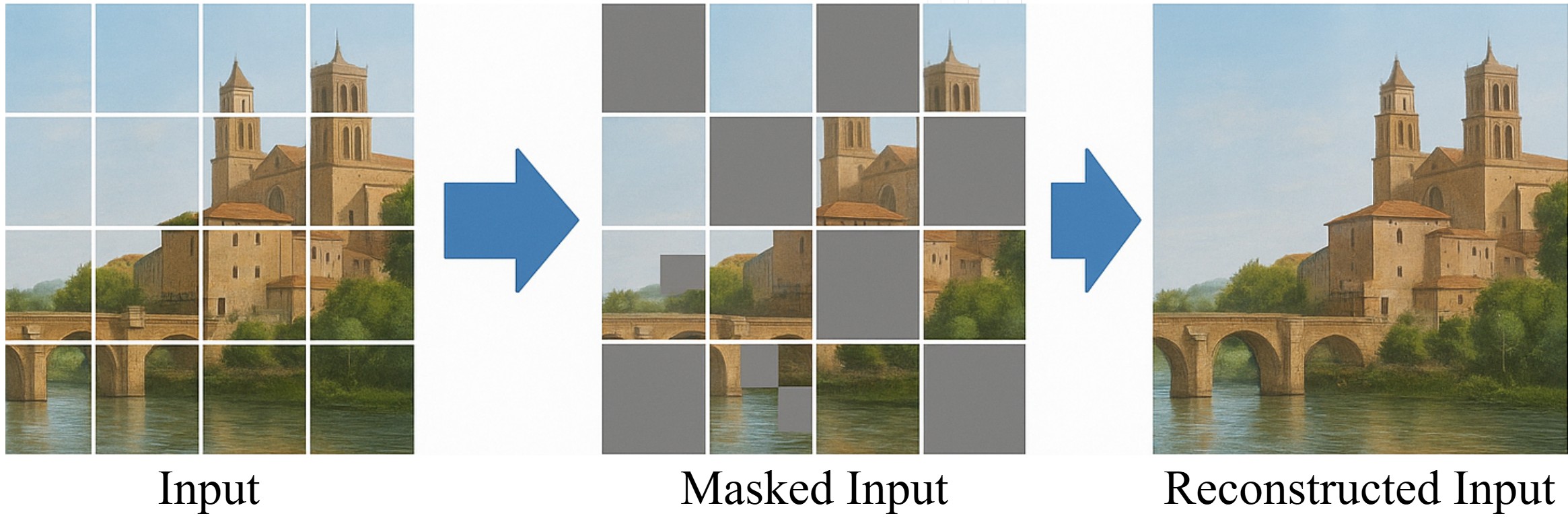}
    \caption{A masked autoencoder learns to reconstruct missing parts of an input by encoding only the unmasked portions and predicting the masked regions.}
    \label{fig:mae}
\end{figure}

Masked Autoencoders \cite{he2022masked} take a different approach to self-supervised learning by masking a significant portion of the input data and training the model to reconstruct the missing parts. He et al. \cite{he2022masked} proposed this method in the context of computer vision, where random patches of an image are removed and the model is trained to reconstruct the missing regions, as illustrated in \autoref{fig:mae}. For example, in their work, Vision Transformers \cite{dosovitskiy2020image} were used to encode only the visible patches, and a lightweight decoder was tasked with reconstructing the masked image. This design encourages the encoder to capture high-level semantic structures and relationships between image regions rather than memorizing local patterns. The effectiveness of MAE lies in its ability to scale to very large datasets without labels while still producing representations that are useful for a range of downstream vision tasks, such as classification and detection. This makes MAE particularly attractive for training foundation models intended to generalize across tasks and data distributions.  

\subsection{\MakeUppercase{Knowledge Distillation}}
As deep learning models grow increasingly large and powerful, deploying them in real-world applications often faces challenges such as high computational cost, limited memory, or the need for real-time inference. Moreover, knowledge acquired by large-scale models during extensive pretraining may not be directly accessible or transferable to smaller models optimized for specific downstream tasks. Knowledge distillation addresses these limitations by providing a mechanism for transferring knowledge from a high-capacity teacher model to a smaller or more specialized student model \cite{hinton2015distilling}.

In its general implementation, knowledge distillation involves training the student model to replicate the outputs or internal behaviors of the teacher \cite{hinton2015distilling}. This can be done by matching the teacher’s predicted probability distributions, aligning intermediate feature representations, or capturing relational structures between data points \cite{gou2021knowledge}. For output-level distillation, the student learns to match the teacher's softened probability distribution using a temperature scaling parameter \((T )\) in the softmax function. The objective is to minimize the Kullback-Leibler (KL) divergence \cite{gou2021knowledge} between the two distributions:

\[
L_{\mathrm{KD}} = \sum_{i} p_i^{(T)} \log\left(\frac{p_i^{(T)}}{q_i^{(T)}}\right)
\]

where \( 
p_i^{(T)} = \frac{\exp(z_i^t / T)}{\sum_j \exp(z_j^t / T)}, \quad
q_i^{(T)} = \frac{\exp(z_i^s / T)}{\sum_j \exp(z_j^s / T)} 
\) are the softened probabilities from the teacher \((z^t)\) and student \((z^s)\) logits, respectively. 

For feature-based distillation, the goal is to align the student's intermediate representations with the teacher's. This is often achieved by minimizing the L2 distance between their feature maps \( (f_t \) and \(f_s)\):

\[
L_{\mathrm{Feat}} = \| f_t(x) - f_s(x) \|_2^2
\]

Relational distillation takes this a step further by transferring the geometric relationships between data points in the feature space. Instead of matching individual features, it captures structural knowledge, for example by penalizing differences in the distances and angles between pairs of data point representations from the teacher and student. The student is optimized to approximate the teacher’s guidance using a combination of a standard supervised loss and one or more of these distillation losses.

Distillation can be categorized into several paradigms based on supervision. In supervised distillation, both teacher and student are trained on labeled data, with the student additionally learning from the soft outputs or features of the teacher \cite{hinton2015distilling}. Semi-supervised distillation leverages unlabeled data by letting the teacher generate pseudo-labels for the student, as seen in Noisy Student \cite{xie2020self}. Self-supervised distillation, on the other hand, relies on pretext tasks without any labels, as exemplified by BYOL \cite{grill2020bootstrap} and DINO \cite{caron2021emerging}, where the student learns to match the features of a momentum-updated teacher through contrastive or alignment-based objectives.

 \subsection{\MakeUppercase{3D Reconstruction}}
3D reconstruction refers to the process of inferring the three-dimensional structure of a scene from two-dimensional observations, such as images or depth maps. It plays a fundamental role in computer vision and graphics, enabling tasks such as scene understanding, augmented reality, and robotic navigation. For foundation models, 3D reconstruction serves as a critical source of spatial and geometric priors, allowing the model to develop a deeper understanding of the physical world and improving its generalization to tasks involving depth, motion, and object permanence. Traditionally, 3D reconstruction has been approached using geometric techniques such as structure-from-motion (SfM), multi-view stereo (MVS), and simultaneous localization and mapping (SLAM) \cite{schonberger2016structure, furukawa2015multi, cadena2016past}. These methods explicitly build representations like point clouds, meshes, or voxel grids by triangulating information across multiple calibrated views, often relying on handcrafted features and optimization-based pipelines.

As data availability and computational capabilities expanded, traditional geometry-based pipelines began to be complemented or supplanted by learning-based approaches that offered greater adaptability and expressive power \cite{choy20163d, eigen2015predicting}. These methods leverage data-driven learning to model complex scene structures and dynamics, reducing the dependence on handcrafted features and rigid geometric assumptions. 

Among these, Neural Radiance Fields (NeRF) \cite{mildenhall2021nerf} represent a transformative approach to 3D scene representation and novel view synthesis, leveraging the power of neural networks to learn continuous volumetric functions from image observations. Unlike traditional 3D reconstruction methods that rely on explicit representations such as point clouds, meshes, or voxels, NeRF adopts an implicit formulation, modeling the scene as a continuous function  \(f_{\theta} : (\mathbf{x}, \mathbf{d}) \rightarrow (\mathbf{c}, \sigma) \),  where \( \mathbf{x} \in \mathbb{R}^3 \) is the spatial location, 
\( \mathbf{d} \in \mathbb{R}^3 \) is the viewing direction, 
\( \mathbf{c} \) is the RGB color, and 
\( \sigma \) is the volumetric density. 
Photorealistic rendering is achieved by integrating color and density along camera rays using the differentiable volume rendering equation \cite{mildenhall2021nerf}: 
\[
C(\mathbf{r}) = \int_{t_n}^{t_f} T(t)\,\sigma(\mathbf{r}(t))\,\mathbf{c}(\mathbf{r}(t), \mathbf{d})\,dt
\]
where \( T(t) = \exp\left( -\int_{t_n}^{t} \sigma(\mathbf{r}(s)) \, ds \right) \) is the accumulated transmittance. This enables high-quality view synthesis from sparse multi-view imagery with known camera poses.

\begin{figure}[htbp]
    \centering
    \includegraphics[width=\columnwidth]{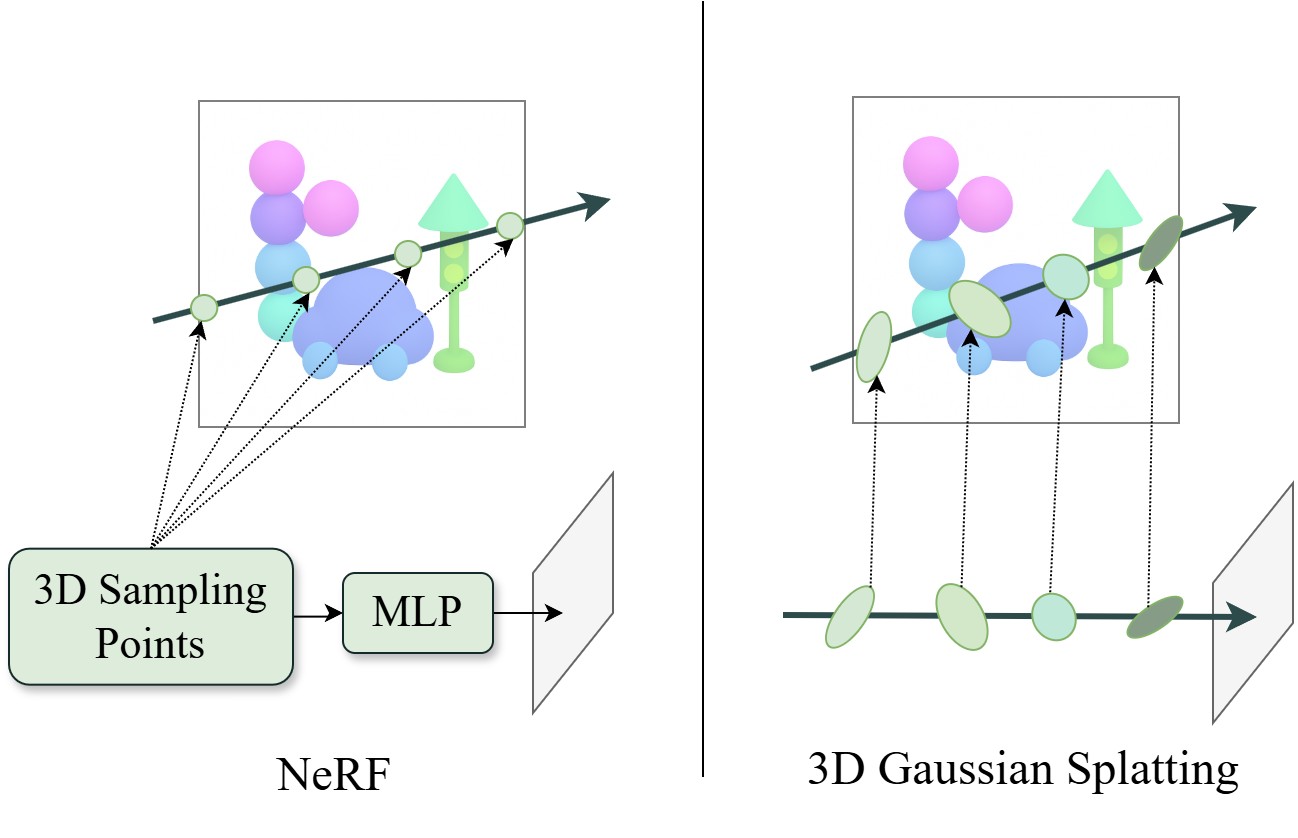}
    \caption{NeRF performs backward rendering by sampling points along each ray and querying an Multi-Layer Perceptron (MLP) to infer color and density, followed by volumetric rendering. In contrast, 3D Gaussian Splatting uses forward rendering by projecting pre-defined 3D Gaussians directly onto the image plane via differentiable Gaussian rasterizer.}
    \label{fig:nerf_vs_gs}
\end{figure}

While NeRF pioneered the use of implicit neural fields for high-fidelity reconstruction, it often incurs significant computational cost due to its dense volumetric sampling and reliance on neural network inference for every query point. In response to these limitations, 3D Gaussian Splatting (3DGS) \cite{kerbl20233d} has emerged as a compelling alternative. Instead of learning an implicit function, 3DGS represents the scene as a set of explicit 3D Gaussians \( \mathcal{G} = \left\{ \left( \boldsymbol{\mu}_i, \boldsymbol{\Sigma}_i, \mathbf{c}_i, \alpha_i \right) \right\}_{i=1}^{N} \), where \( \boldsymbol{\mu}_i \in \mathbb{R}^3 \) denotes the mean position, \( \boldsymbol{\Sigma}_i \in \mathbb{R}^{3 \times 3} \) is the covariance matrix encoding anisotropic shape, \( \mathbf{c}_i \in \mathbb{R}^3 \) is the RGB color, and \( \alpha_i \in [0, 1] \) is the opacity. Each Gaussian is projected onto the image plane via a differentiable forward rasterizer, and their contributions are accumulated using alpha compositing.\cite{kerbl20233d} Unlike NeRF, which relies on dense sampling and neural inference, 3DGS directly projects these Gaussians into the image plane and composites them using a differentiable rendering pipeline. This enables photorealistic rendering at significantly higher speeds, making 3DGS particularly well-suited for real-time applications. \autoref{fig:nerf_vs_gs} illustrates the difference in rendering method between NeRF and 3D gaussian splatting.

\begin{table*}[t]
  \centering
  \caption{Comprehensive Comparison between Traditional Deep Learning and Foundation Models}
  \label{tab:ai_paradigm_transposed_grid}
  \begin{tabular}{
      >{\centering\arraybackslash}p{0.20\linewidth}|
      >{\centering\arraybackslash}p{0.18\linewidth}|
      >{\centering\arraybackslash}p{0.18\linewidth}|
      >{\centering\arraybackslash}p{0.18\linewidth}|
      >{\centering\arraybackslash}p{0.18\linewidth}}
    \hline
    \textbf{Attribute} & \textbf{Traditional Deep Learning} & \textbf{Large Language Models (LLMs)} & \textbf{Vision Foundation Models (VFMs)} & \textbf{Vision-Language Models (VLMs)} \\
    \hline
    \textbf{Primary Modality} & Vision (typically) & Text & Vision (Images/Video) & Vision + Text \\
    \hline
    \textbf{Capability} & Performs a single, pre-defined task (e.g., detect cars). & Understands, generates, and reasons with human language. & Creates general-purpose visual representations for any-object perception. & Aligns visual content with language for cross-modal understanding. \\
    \hline
    \textbf{Training Approach} & Supervised training from scratch on a task-specific, closed-set dataset. & Self-supervised pre-training on vast text datasets. & Self-supervised pre-training on broad, diverse visual datasets. & Pre-training on large-scale paired image-text data. \\
    \hline
    \textbf{Example Models} & YOLO v3 \cite{redmon2018yolov3}, ResNet \cite{he2016deep} & GPT-series \cite{achiam2023gpt}, BERT \cite{devlin2019bert} & SAM \cite{kirillov2023segment}, DINO v2 \cite{oquab2023dinov2} & CLIP \cite{radford2021learning}, Grounding DINO \cite{liu2024grounding} \\
    \hline
    \textbf{Generalization} & Low & High & High & High \\
    \hline
    \textbf{Specialization} & High & Low & Medium & Low \\
    \hline
    \textbf{Computational Cost} & Low & High & High & High \\
    \hline
  \end{tabular}
\end{table*}

  \subsection{\MakeUppercase{Diffusion}}
Denoising diffusion probabilistic models are a powerful class of generative models that have demonstrated exceptional capability in producing high-fidelity images, videos, and 3D structures \cite{ho2020denoising}. These models operate by learning to reverse a predefined noising process, where Gaussian noise is incrementally added to the data over a series of time steps. During training, the model learns how to denoise this corrupted data step-by-step, ultimately acquiring the ability to generate realistic samples from pure noise during inference. The structured and progressive nature of this generative process allows diffusion models to be adapted for a wide range of practical applications. For example, in text-to-image generation, models like Stable Diffusion are conditioned on natural language descriptions and generate images that faithfully align with textual prompts \cite{rombach2022high}. In video generation, diffusion models have been used to synthesize temporally coherent sequences from single frames or motion cues \cite{ho2022video}. In 3D generation, they can be combined with neural fields or voxel grids to produce realistic object shapes or scenes \cite{zhou20213d, shim2023diffusion}. These examples demonstrate the versatility of diffusion models in capturing complex data distributions, which is critical for the general-purpose nature of foundation models. For foundation models, diffusion represents a robust mechanism for learning general-purpose priors that span modalities, offering scalability, compositionality, and interpretability—key attributes for building flexible and reusable generative backbones.

  \subsection{\MakeUppercase{General Foundation Models}}
  General foundation models refer to large-scale neural networks trained on broad and diverse datasets with the goal of acquiring versatile and reusable representations across a wide range of downstream tasks \cite{bommasani2021opportunities}. These models are typically pre-trained using self-supervised or weakly supervised objectives and are designed to serve as a universal backbone that can be adapted to different modalities, including vision, language, audio, and 3D data. The central idea is that by leveraging massive data and model scale, a single foundation model can internalize common structures and semantics that generalize well across tasks and domains, reducing the need for task-specific training from scratch. This universality stands in contrast to traditional deep learning models, which are highly specialized for a single task, offering high predictability but limited adaptability to novel scenarios. This makes foundation models an attractive solution in AI systems that demand flexibility, scalability, and rapid adaptation, such as in perception, reasoning, and planning modules for robotics and autonomous driving.

Vision foundation models are a class of large-scale neural networks trained on diverse and extensive visual datasets with the objective of acquiring general-purpose visual representations. These models are designed to perform a wide range of visual tasks—including classification, segmentation, and object detection—without requiring task-specific re-training. A key strength of vision foundation models lies in their ability to generalize to unseen categories and tasks through zero-shot or few-shot learning. For instance, models like the Segment Anything Model (SAM) \cite{kirillov2023segment} leverage prompt-based interfaces to segment novel objects with minimal supervision, making them highly adaptable and label-efficient. Similarly, models such as DINO \cite{caron2021emerging, oquab2023dinov2} learn semantic-rich representations by exploiting self-supervised objectives, enabling robust performance across downstream vision benchmarks.

Large Language Models (LLMs) are a class of transformer-based neural networks trained on massive corpora of text data with the goal of learning broad linguistic knowledge and reasoning capabilities. These models are capable of understanding and generating human language across diverse domains, and are typically pre-trained using next-token prediction or masked language modeling \cite{brown2020language, devlin2019bert}. A distinguishing feature of LLMs is their ability to perform zero-shot and few-shot generalization, making them highly effective at a wide array of language tasks such as question answering, summarization, code generation, and dialogue \cite{brown2020language, achiam2023gpt}. Therefore, They become foundational components for building intelligent systems that require natural language understanding and generation.
\begin{figure*}[htbp]
    \centering
    \includegraphics[width=\textwidth]{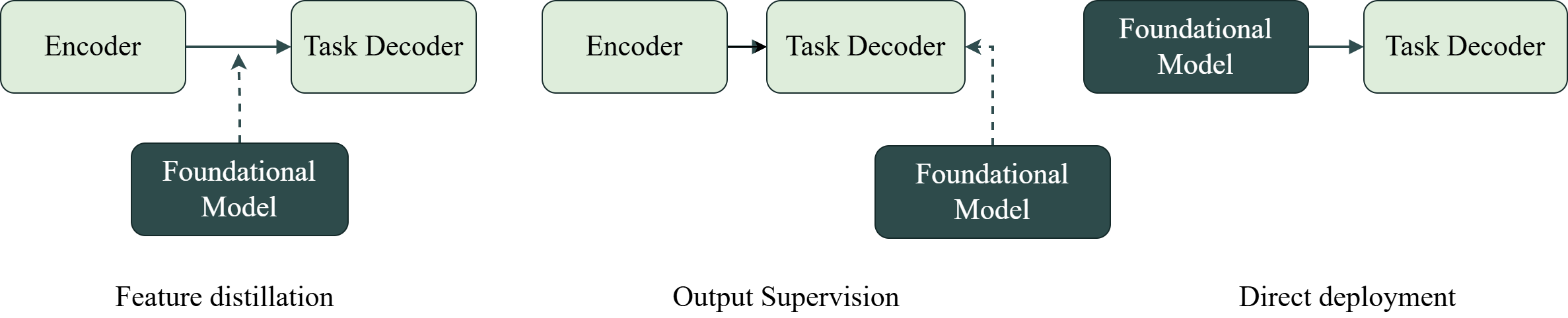}
    \caption{From left to right: (1) \textbf{Feature distillation}, where features from a foundation model supervise a encoder; (2) \textbf{Output supervision}, where the foundation model provides pseudo labels to guide training; and (3) \textbf{Direct deployment}, where the foundation model is used directly in the perception pipeline to produce task-specific outputs. }
    \label{fig:generalizedknowledge}
\end{figure*}

Building on the strengths of both domains, Vision Language Models (VLMs) integrate the capabilities of vision foundation models and large language models by learning joint representations that align visual content with textual semantics. These models are trained on paired image-text data and are capable of understanding and reasoning across modalities, allowing them to associate visual elements with descriptive language in an open-vocabulary manner. For instance, CLIP (Contrastive Language–Image Pre-training) \cite{radford2021learning} learns to match images and text by maximizing similarity between paired inputs while minimizing similarity between mismatched pairs. This enables zero-shot classification by comparing an image's embedding to textual descriptions of possible categories. Grounding DINO \cite{liu2024grounding} builds upon this concept by extending it to structured vision tasks such as object detection and segmentation, leveraging language prompts to identify and localize objects in complex scenes. These models expand the expressiveness and generalization potential of perception systems, particularly in domains like autonomous driving where flexible, language-aware interpretation of scenes can enhance robustness and interactivity.

To summarize the distinctions, the primary advantage of foundation models over traditional deep learning for autonomous driving perception lies in their superior generalization and adaptability. While traditional models offer high predictability and lower computational cost, their performance is confined to pre-defined tasks. In contrast, foundation models operate on a different paradigm. LLMs provide high-level reasoning on text, VFMs create general-purpose visual representations for any-object perception, and VLMs align vision with language for unparalleled cross-modal understanding. This trade-off—accepting higher computational costs for a massive leap in generalization—is what makes foundation models a transformative approach for handling the unpredictable nature of real-world driving scenarios. Table \ref{tab:ai_paradigm_transposed_grid} provides a direct comparison of these models.

\section{\MakeUppercase{Generalized Knowledge in perception models}}
 In the context of autonomous driving perception, the general foundation models provide robust visual, semantic, and reasoning knowledge encoding implicit world knowledge. By leveraging cross-domain insights, perception systems can generalize more effectively to new environments, improve robustness under varying conditions, and reduce reliance on extensive, costly labeled driving data.

To integrate these foundation models into the autonomous driving stack, three primary mechanisms have emerged. The first approach, feature-level distillation, uses outputs or intermediate features from a foundation model to supervise a smaller, task-specific perception network. This method results in compact models that inherit knowledge from the foundation model while maintaining high inference efficiency. The second approach, pseudo label supervision, utilizes foundation models to generate automatically derived annotations such as segmentation masks or bounding boxes. These pseudo labels enable scalable, label-efficient training of perception networks, reducing the need for costly manual labeling and facilitating adaptation to diverse domains. The third approach, direct integration, incorporates the foundation model itself within the perception pipeline, serving as a feature extractor or a modular component that provides rich, pre-learned representations. This method allows direct use of powerful pretrained features without requiring retraining from scratch. Collectively, these mechanisms enable the adaptation of general-purpose models to the specialized, safety-critical demands of automotive perception. \autoref{fig:generalizedknowledge} illustrates the three different approaches commonly used to integrate foundation models into the autonomous driving stack.

\begin{table*}[t]
  \centering
  \caption{Representative General-Knowledge Adaptation Methods Leveraging Vision Foundation Models and Vision Language Models}
  \label{gm-methods}
  \begin{tabular}{>{\centering\arraybackslash}p{0.13\linewidth}|
                  >{\centering\arraybackslash}p{0.18\linewidth}|
                  >{\centering\arraybackslash}p{0.10\linewidth}|
                  >{\centering\arraybackslash}p{0.17\linewidth}|
                  >{\centering\arraybackslash}p{0.32\linewidth}}
    \hline
    \textbf{Method} & \textbf{Adaptation Strategy} & 
    \textbf{Foundation Model Type} &
    \textbf{Foundation Model(s)} & 
    \textbf{Downstream Task}\\
    \hline
    SLidR \cite{sautier2022image} & Feature-level distillation & VFM & DINO & LiDAR model pretraining \\
    \hline
    SEAL  \cite{liu2023segment} & Feature-level distillation & VFM & SAM, SEEM & LiDAR model pretraining \\
    \hline
    SAL \cite{ovsep2024better} & Pseudo-label supervision & VFM, VLM & SAM, CLIP & Zero-shot 3D semantic \& instance segmentation \\
    \hline
    SAM4UDASS \cite{10366854} & Pseudo-label supervision & VFM & SAM & 2D semantic segmentation with domain shift \\
    \hline
    OccNeRF  \cite{zhang2023occnerf} & Pseudo-label supervision & VFM, VLM & SAM, Grounding DINO & Camera-only 3D occupancy \\
    \hline
    Open3DWorld \cite{cheng2024open} & Feature-level distillation & VLM & CLIP (text encoder) & LiDAR open-vocabulary detection \\
    \hline
    OVO  \cite{tan2023ovo} & Feature-level distillation & VLM & LSeg & Camera-only open-vocabulary 3D occupancy \\
    \hline
    CLIP2Scene \cite{chen2023clip2scene} & Feature-level distillation & VLM & MaskCLIP & LiDAR 3D semantic segmentation \\
    \hline
    VLM2Scene \cite{liao2024vlm2scene} & Feature-level distillation & VFM, VLM & SAM, BLIP-2, CLIP & LiDAR model pretraining \\
    \hline
    UP-VL \cite{najibi2023unsupervised} & Feature-level distillation / pseudo-label supervision & VLM & OpenSeg & LiDAR open-vocabulary detection \\
    \hline
    OpenSight \cite{zhang2024opensight} & Feature-level distillation / pseudo-label supervision & VLM & Grounding DINO & LiDAR open-vocabulary detection \\
    \hline
  \end{tabular}
  \label{tab:general_knowledge_methods}
\end{table*}

  \subsection{\MakeUppercase{Vision Foundation Model}}
  Vision Foundation Models (VFMs) are being explored for their potential to enhance autonomous driving perception. Since VFMs are essentially vision models, they can be directly integrated into not only 2D image pipelines but also extended to 3D perception pipelines with appropriate modifications, such as projecting image features into 3D space or aligning them with point cloud data \cite{zhang2023sam3d}. This versatility allows VFMs to contribute high-level semantic understanding across modalities. However, direct integration is often limited in practice due to high computational cost and latency, especially for real-time applications. As a result, VFMs are more commonly employed as a source of rich features for distillation into lighter models or as generators of pseudo labels to bootstrap training without manual annotations. 
  
\subsubsection{Feature-level Distillation}
VFM distillation often targets 3D LiDAR-based models, leveraging 2D image-based VFMs due to LiDAR’s data scarcity. SLidR \cite{sautier2022image} distills a 2D VFM like DINO into a 3D LiDAR network via contrastive learning, aligning features from image superpixels and corresponding LiDAR points. This cross-modal distillation allows the 3D model to inherit the semantic richness of 2D representations without requiring dense 3D labels. Subsequent improvements have been made by SEAL \cite{liu2023segment}, which enhances this method by utilizing segmentation-based VFM like SAM and SEEM to generate high-quality semantic superpixels instead of clustering pixels based on low-level visual similarity. It further introduces a semantic-guided contrastive distillation framework that selectively matches semantically similar regions across modalities, reducing noisy supervision and improving feature transfer.

\subsubsection{Pseudo Label Supervision}
VFMs can also generate pseudo labels, which not only mitigates the need for costly manual labeling, but also enables the transfer of general visual knowledge encoded in large-scale VLMs into task-specific perception models, thereby enhancing their semantic understanding and generalization capabilities. For instance, SAL \cite{ovsep2024better} employs SAM to generate 2D instance masks by projecting them onto the 3D point cloud via geometric calibration. These masks serve as supervisory signals for training a 3D instance segmentation model, enabling the transfer of detailed object-level information from image to point cloud without requiring manual 3D annotations. Additionally, SAM4UDASS \cite{10366854} leverages SAM-generated masks to improve pseudo label quality in an unsupervised domain adaptation setting. It selectively integrates high-confidence regions from SAM outputs to refine segmentation predictions, particularly benefiting small and rare classes where conventional pseudo labeling tends to be less reliable. OccNeRF \cite{zhang2023occnerf} adopts a multi-source supervision strategy by combining image-derived depth maps and SAM-based segmentations to guide the learning of 3D occupancy fields in LiDAR-free scenes. This hybrid approach enables geometry-aware scene understanding using only RGB inputs, demonstrating the versatility of VFM-based pseudo labeling as a complementary method alongside traditional LiDAR-based approaches. While LiDAR offers precise depth sensing, VFM-derived labels from RGB inputs provide rich semantic information without requiring expensive 3D annotations. 

  \subsection{\MakeUppercase{Vision Language Model }}
Vision-Language Models (VLMs) provide a powerful way to incorporate high-level semantic understanding from language into visual perception models, which is especially valuable in open-world scenarios encountered in autonomous driving. Similar to VFMs, the focus for the applications of VLM also lies in distillation and pseudo-labeling. 
    
\subsubsection{Feature-level Distillation}
VLMs like CLIP typically contain two components: a text encoder and an image encoder, each responsible for extracting features from their respective modalities \cite{radford2021learning}. This dual-encoder setup provides the flexibility to use either modality independently or in combination, which proves particularly valuable in autonomous driving scenarios where both textual cues and visual content can complement each other to enhance perception.

Focusing on the text modality, Open3DWorld \cite{cheng2024open} uses only the text encoder part of CLIP to enhance 3D object detection in LiDAR point clouds. Specifically, it transforms object category labels into text embeddings using the CLIP text encoder and aligns these with LiDAR bird’s-eye view (BEV) features through a cross-modal matching mechanism. This design allows the model to incorporate rich semantic priors from large-scale vision-language pretraining into 3D detection, enabling it to better generalize to diverse object categories and open-set scenarios without relying solely on point cloud features.

 Turning to the image modality, VLM distillation faces a gap: models like CLIP, trained on image-text pairs for global classification, capture coarse semantics but miss fine local details. Autonomous driving tasks demand dense, pixel- or object-level supervision, so researchers prefer VLMs enriched with local information—either fine-tuned for dense prediction or extended with pixel-wise supervision. Such models yield features better aligned with the spatial and semantic resolution needed for complex driving perception.

For example, for semantic segmentation, OVO \cite{tan2023ovo} distills the image features from LSeg—an open-vocabulary segmentation model trained with pixel-level supervision and text supervision—into a 3D occupancy network. OVO introduces a cross-modal distillation pipeline that transfers semantic knowledge from LSeg’s 2D predictions to 3D voxels using multi-view consistency and voxel-wise contrastive learning. This enables the model to generate semantic occupancy grids for arbitrary open-vocabulary classes without relying on 3D ground-truth labels. Similarly, CLIP2Scene \cite{chen2023clip2scene} distills both the text and image features from MaskCLIP, which leverages masked self-distillation to retain local semantic structure. It aligns these features with point cloud representations using a contrastive loss, enabling annotation-free point cloud semantic segmentation. CLIP2Scene also introduces a selective positive sampling strategy to enhance the discriminability of 3D features by aligning them with localized language-guided semantics. 

While the above examples use VLMs tailored for local semantics, traditional CLIP—focused on global image-level features—can still be effective when paired with models providing localized cues. For instance, VLM2Scene \cite{liao2024vlm2scene} introduces a region-aware pipeline: SAM generates fine-grained masks, BLIP-2 captions each region, and CLIP encodes the captions into localized embeddings. Visual features from the masks are aligned with these embeddings via a regional contrastive loss, preserving spatial granularity and semantic diversity. This approach compensates for CLIP's global-only training and improves fine-grained perception in open-world driving.

\subsubsection{Pseudo Label Supervision}
Similar to VFMs, VLMs can also serve a dual purpose in object detection tasks—not only as a source of feature distillation, but also as a generator of pseudo ground truth boxes. VLMs can provide semantic supervision from large-scale image-text pretraining while simultaneously generating auxiliary labels to train downstream models in a scalable, label-efficient manner. UP-VL \cite{najibi2023unsupervised} proposes a unified point-vision-language framework that leverages OpenSeg for extracting dense semantic features, which are then distilled into a 3D detector network. In addition to feature distillation, UP-VL utilizes the textual outputs of OpenSeg in conjunction with spatiotemporal clustering and filtering algorithms to automatically generate 3D bounding boxes and object tracklets, thereby producing high-quality pseudo labels without requiring human annotations. These pseudo labels serve as supervisory signals for training 3D detectors in a scalable way. Similarly, OpenSight \cite{zhang2024opensight} also performs cross-modal distillation by aligning 2D visual-language features with 3D point cloud representations. However, instead of generating pseudo boxes, it integrates Grounding DINO as a teacher to directly supervise 3D object detection. This bypasses the need for heuristic box proposals and allows OpenSight to inherit object grounding capabilities from large-scale open-world training. 

Table \ref{gm-methods} overviews the main ways recent work repurposes Vision Foundation Models and Vision Language Models for autonomous‑driving perception mentioned in the above sections.

\begin{table*}[t]
  \centering
  \caption{Representative Perception-Reasoning Integration Strategies in LLM-based Driving Systems}
  \label{tab:llm_methods}
  \begin{tabular}{>{\centering\arraybackslash}p{0.15\linewidth}|
                  >{\centering\arraybackslash}p{0.2\linewidth}|
                  >{\centering\arraybackslash}p{0.25\linewidth}|
                  >{\centering\arraybackslash}p{0.3\linewidth}}
    \hline
    \textbf{Representative Method} & 
    \textbf{Model Input} & 
    \textbf{Perception-Reasoning Supervision During Training} & 
    \textbf{Detail Level of Perception Available to Planner} \\
    \hline
    GPT-Drive \cite{mao2023gpt} & Textual summaries from separate perception stack & N/A (perception outside LLM) & High-level symbolic \\
    \hline
    OmniDrive \cite{wang2024omnidrive} & Raw camera images & No & Mid-level geometry-aware tokens \\
    \hline
    DriveGPT4 \cite{xu2024drivegpt4} & Raw camera images & Yes (caption-based visual instruction tuning) & Coarse semantic  scene understanding \\
    \hline
    Dolphins \cite{ma2024dolphins} & Raw camera images & Yes (QA on objects \& relations) & Coarse semantic  scene understanding \\
    \hline
    EMMA \cite{hwang2024emma} & Raw camera images & Yes (joint 3D detection and planning losses) & Fine-grained metric 3D boxes and trajectories \\
    \hline
  \end{tabular}
\end{table*}

\subsection{\MakeUppercase{Large Language Model}}

 Due to the strong capabilities in abstraction, reasoning, and instruction following, large language models (LLMs) have recently gained traction in the context of autonomous driving perception. Their potential lies in unifying multi-modal sensor data and complex decision-making processes within a language-based framework. At an implementation level, this is often achieved by first processing raw sensor data through a specialized encoder to extract feature vectors. These vectors are then projected into the LLM's word embedding space, effectively creating sensor tokens that can be processed alongside text. This allows models like Lidar-LLM \cite{yang2025lidar} to use linguistic prompts to align, explain, and interpret fused sensor data, enhancing system flexibility and interpretability.

Since it is challenging to distill language features, which are often abstract, context-dependent, and distributed, into traditional perception models that expect structured and localized representations, recent approaches increasingly opt to directly integrate LLMs into the full autonomous driving pipeline. The core implementation strategy is to reframe the driving task as a sequential language modeling problem. This involves creating a unified vocabulary of tokens that represent not only words but also discretized sensor measurements and action commands. The LLM is then trained to predict the next token in a sequence, which could be a word for an explanation or an action token for vehicle control.

Traditionally, LLMs have been employed to generate planning outcomes based on given perception data input obtained through conventional methods, such as GPT-Drive \cite{mao2023gpt}. GPT-Drive relies on perception results that are converted into textual descriptions summarizing scene context, such as detected objects and road layouts, and formulates planning as a language modeling problem, using textual prompts to steer the driving behavior. Building upon this foundation, OmniDrive \cite{wang2024omnidrive} introduces a more tightly coupled framework where the LLM is directly conditioned on multi-view image features and 3D spatial cues, allowing it to reason jointly over perception and planning. In OmniDrive, the model leverages multi-view visual encodings and 3D-aware language prompts to enable fine-grained scene understanding that guides downstream planning decisions. This allows the planner to utilize more low-level perception information instead of directly consuming structured, high-level perception outputs.

More recent approaches have embraced end-to-end pipelines where LLMs directly ingest minimally processed sensor inputs to derive planning decisions. While intermediate perception outputs are not strictly required, incorporating perception-related tasks during training has been shown to enhance reasoning. For instance, DriveGPT4 \cite{xu2024drivegpt4} is fine-tuned on visual-textual pairs, pairing camera inputs with human-annotated captions of driving-relevant semantics. Dolphins \cite{ma2024dolphins}, in contrast, uses a chain-of-thought prompting mechanism and a perception pre-training phase where the LLM learns to answer questions about objects and spatial relations. These perception-focused tasks enrich the model’s internal representations, improving its capacity for context-aware planning.

This progression sets the stage for models that integrate fine-grained perception into the LLM framework. For example, EMMA \cite{hwang2024emma} extends LLM-based perception by incorporating dense 3D object detection capabilities directly into the language model pipeline. It reformulates traditional 3D perception tasks as a series of multimodal question-answering prompts, allowing the LLM to predict object categories, 3D bounding boxes, and spatial relations from encoded sensor inputs. This enables the model not only to detect and localize objects with high precision but also to output trajectory coordinates with enhanced geometric consistency. As a result, EMMA demonstrates state-of-the-art performance in 3D object detection and significantly outperforms existing approaches in downstream motion planning tasks.

Table \ref{tab:llm_methods} summarizes how perception is incorporated into the LLM-based driving systems covered in this section.

\subsection{\MakeUppercase{Key Challenges in Generalized Knowledge}}
Each model family—Vision Foundation Models (VFMs), Vision-Language Models (VLMs), and Large Language Models (LLMs)—presents a distinct set of advantages and inherent limitations. VFMs excel at learning dense visual semantics from large-scale image pre-training but struggle with the domain gap to specialized sensor data and often lack explicit 3D geometric understanding from their 2D-centric training. VLMs offer powerful open-vocabulary recognition, allowing for flexible supervision, yet this semantic strength often comes at the cost of weak geometric detail and a reliance on empirical prompt engineering. LLMs provide unprecedented capabilities for unified reasoning and human-interpretable explanations, but their outputs can lack precise pixel-level grounding and are susceptible to dangerous hallucinations. These trade-offs, summarized in Table \ref{tab:gk_pros_cons}, are crucial for understanding the current research landscape and the specific hurdles that must be overcome. The following paragraphs elaborate on the most pressing of these cross-cutting challenges.

\textbf{Domain Gap}: A primary challenge lies in bridging the gap between the generalized knowledge acquired by foundational models during pre-training, typically on web-scale text and 2D image data, and the specific requirements of autonomous driving perception. Integrating specialized autonomous driving sensors, such as LiDAR and Radar, which provide crucial 3D geometric and velocity information, proves difficult. These sensors generate data modalities significantly different from the web-based data foundational models are often trained on, and the scale of available, annotated autonomous driving sensor data does not match the internet-scale corpora used for pre-training. This disparity hinders the direct application or fine-tuning of existing foundational models for optimal utilization of these critical sensor streams, necessitating advancements in cross-modal learning and fusion architectures. Therefore, it is crucial to advance domain adaptation methods that more effectively bridge the distributional gap between pretraining datasets and autonomous driving sensor data. These methods should aim not only to align low-level features, but also to reduce high-level semantic mismatches that often lead to degraded performance or unsafe predictions in deployment scenarios.

\textbf{Hallucination Risk}: The phenomenon of hallucination—where models produce outputs inconsistent with the provided sensor reality—poses a severe safety risk \cite{huang2025survey}. Such deviations from ground truth can lead to catastrophic failures, making reliability a paramount concern. A deeper investigation into the root causes of such failures—whether they stem from data biases, architectural limitations, or reasoning flaws—is therefore necessary to guide the development of robust countermeasures. To mitigate this, research is moving beyond mere detection toward active mitigation strategies. One promising direction is grounding the model's outputs in external, verifiable information. Techniques like Retrieval-Augmented Generation \cite{lewis2020retrieval} are being explored to force the model to base its reasoning on a trusted knowledge base, such as HD maps or a library of traffic laws, rather than generating unverified assertions. Another approach involves architectural improvements for internal validation, such as self-critique mechanisms where a model is trained to question its own outputs for physical plausibility and logical consistency \cite{asai2024self}. The development of rigorous benchmarks remains critically important, not only for detection but for evaluating the efficacy of these emerging mitigation techniques in complex and adversarial scenarios.

\textbf{Latency and Efficiency}: Significant practical hurdles exist in deploying foundation models within the stringent constraints of an autonomous vehicle's hardware and software stack. The immense size and computational cost of these models present a direct conflict with the real-time processing requirements of autonomous driving, where perception-to-action latencies must be on the order of milliseconds. The inference time for a large foundation model can easily exceed this budget, rendering its outputs obsolete for immediate vehicle control. This necessitates substantial research into model optimization techniques such as quantization, pruning, and knowledge distillation to create smaller, more efficient variants without catastrophically degrading performance. Furthermore, integrating these models into existing, highly-optimized autonomous vehicle software pipelines is a non-trivial engineering task, requiring careful consideration of data flow, hardware acceleration, and the overall system architecture to ensure safety and reliability.

\textbf{Interpretability}:  The lack of interpretability in large perception foundation models presents a critical barrier to their safe and trustworthy deployment in autonomous driving systems. To address this challenge, researchers are increasingly exploring natural language and multimodal explanation techniques that aim to translate complex AI decisions into human-readable narratives. Effective explainability mechanisms can support debugging, ensure compliance with safety regulations, and foster both user and regulatory trust. A recent example of progress is Driving with LLMs \cite{chen2024driving}, which enhances interpretability by generating natural language explanations from structured scene representations. This work reflects a broader trend toward language-driven, interpretable AI systems for autonomous driving.

\begin{table*}[t]
  \centering
  \caption{Comparison of Foundation Model Families: Strengths and Open Challenges}
  \label{tab:gk_pros_cons}
  \begin{tabular}{>{\raggedright\arraybackslash}p{0.25\linewidth}|>{\raggedright\arraybackslash}p{0.35\linewidth}|>{\raggedright\arraybackslash}p{0.35\linewidth}}
    \hline
    \textbf{Model Family} & \textbf{Key Strengths} & \textbf{Key Limitations / Open Challenges} \\
    \hline
    Vision Foundation Models (VFM) 
      & Dense visual semantics from large-scale image pre-training; Lifts LiDAR models without extra 3D labels
      & 2D-only training lacks explicit 3D geometry; Domain gap in night / adverse-weather / fisheye settings \\
    \hline
    Vision–Language Models (VLM) 
      & Open-vocabulary recognition of unseen classes; Flexible supervision via joint visual–semantic embedding 
      & Strong semantic information but weak geometric detail; Empirical prompt engineering \& domain adaptation \\
    \hline
    Large Language Models (LLM) 
      & Unified reasoning over multimodal inputs and planning; Human-interpretable chain-of-thought explanations
      & Weak pixel-level grounding; geometry approximate, Safety, verifiability, and hallucination remain open issues \\
    \hline
  \end{tabular}
\end{table*}

\section{\MakeUppercase{Spatial understanding}}
Spatial understanding enables autonomous vehicles to build coherent 3D representations of their environment, capturing object identity, geometry, and context. Traditional systems rely on discrete detections and hand-crafted pipelines, which struggle with irregular or unfamiliar scenes. Spatially-aware models offer a more holistic view by embedding fine-grained geometric and semantic understanding. 

A key benefit of spatial understanding is its ability to support holistic scene understanding by capturing detailed geometric and contextual information. By reasoning over the spatial relationships between scene elements, autonomous systems can interpret complex settings, distinguish between drivable and non-drivable areas, and identify obstacles, even when they do not match predefined object categories \cite{xu2025survey}. This is especially important in diverse and unstructured environments where reasoning about free space, occlusions, and scene layout is critical for safe navigation. Spatial understanding also enables generalization and task unification. A robust spatial representation can serve as a shared foundation for multiple downstream tasks, including object detection, semantic segmentation, motion prediction, and planning. A unified model allows for consistent reasoning across these tasks, improving system performance and eliminating redundant computations \cite{huang2024neural}.  Table \ref{spatial-methods} summarizes each method’s core design choices—strategy, input sensors, output representation, and supervision.

\subsection{\MakeUppercase{Volumetric Models}}
Recent advancements in autonomous driving perception have increasingly focused on the development of continuous volumetric representations to achieve a more comprehensive spatial understanding of the environment. This marks a departure from traditional systems that often rely on discrete, object-level predictions like bounding boxes or semantic segmentations. Volumetric models, inspired by computer graphics techniques such as volume rendering and neural representations, aim to encode a dense and holistic understanding of the driving scene. By interpreting sensor inputs—primarily camera images and LiDAR point clouds—as projections of an underlying three-dimensional reality, these methods enable foundational models to reason about spatial occupancy, semantics, and geometry in a unified framework. 
   
\subsubsection{3D Learning via Explicit Occupancy Networks}

\begin{figure}[htbp]
    \centering
    \includegraphics[width=\columnwidth]{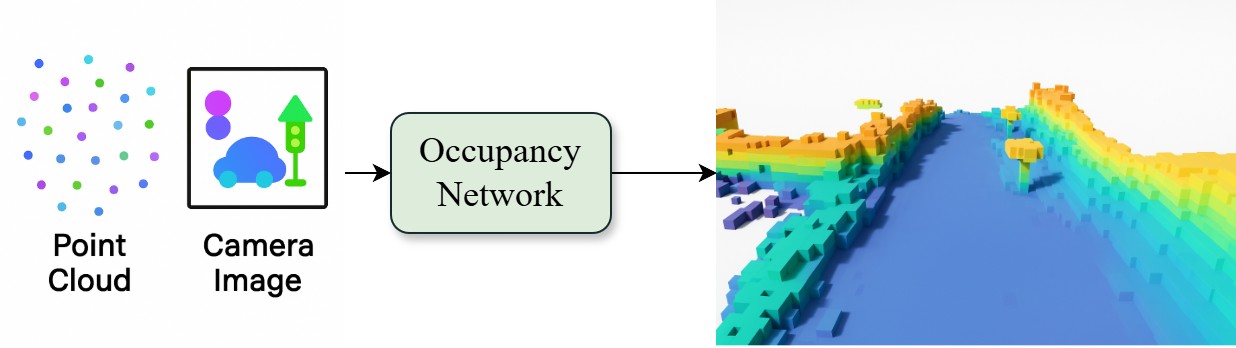}
    \caption{ Illustration showing the modeling of input data into an occupancy grid representation, providing the model with a spatial understanding of the environment.}
    \label{fig:occupancy}
\end{figure}
A prominent approach within volumetric modeling involves Occupancy Networks, which discretize the space surrounding the ego-vehicle into a voxel grid. These models predict the occupancy status and semantic class for each voxel, offering an explicit 3D representation derived conceptually from classical Occupancy grid mapping. This explicit spatial encoding captures complex environmental structures often lost in 2D projections, providing a crucial foundation for downstream tasks such as 3D object detection, semantic segmentation, and motion planning. \autoref{fig:occupancy} illustrates the use of explicit 3D occupancy grid to learn a 3D representation. 

Several notable architectures have emerged to implement 3D occupancy modeling within neural networks, employing diverse strategies for representation and reasoning. For instance, TPVFormer \cite{huang2023tri} constructs a tri-perspective view by orthogonally projecting multi-view image features onto bird's-eye, frontal, and lateral planes, utilizing a transformer to aggregate these multi-planar spatial features into a cohesive volumetric understanding. Voxformer \cite{li2023voxformer} adopts a two-stage pipeline for monocular semantic scene completion, first predicting depth to generate sparse voxel proposals and subsequently refining these proposals through transformer-based volumetric reasoning. Building on this, OccFormer \cite{zhang2023occformer} employs a dual-path transformer to separate local and global spatial encoding within the horizontal voxel plane, incorporating techniques like preserve-pooling and class-aware sampling to enhance feature fidelity. Addressing computational efficiency, SparseOcc \cite{liu2024fully} proposes a sparse query-based network that operates on selectively sampled voxel tokens guided by 2D segmentation cues, thereby enabling efficient volumetric inference without requiring dense 3D computations.

\subsubsection{3D Learning via 2D Supervision with Neural Rendering}
 
\begin{figure*}[htbp]
    \centering
    \includegraphics[width=\textwidth]{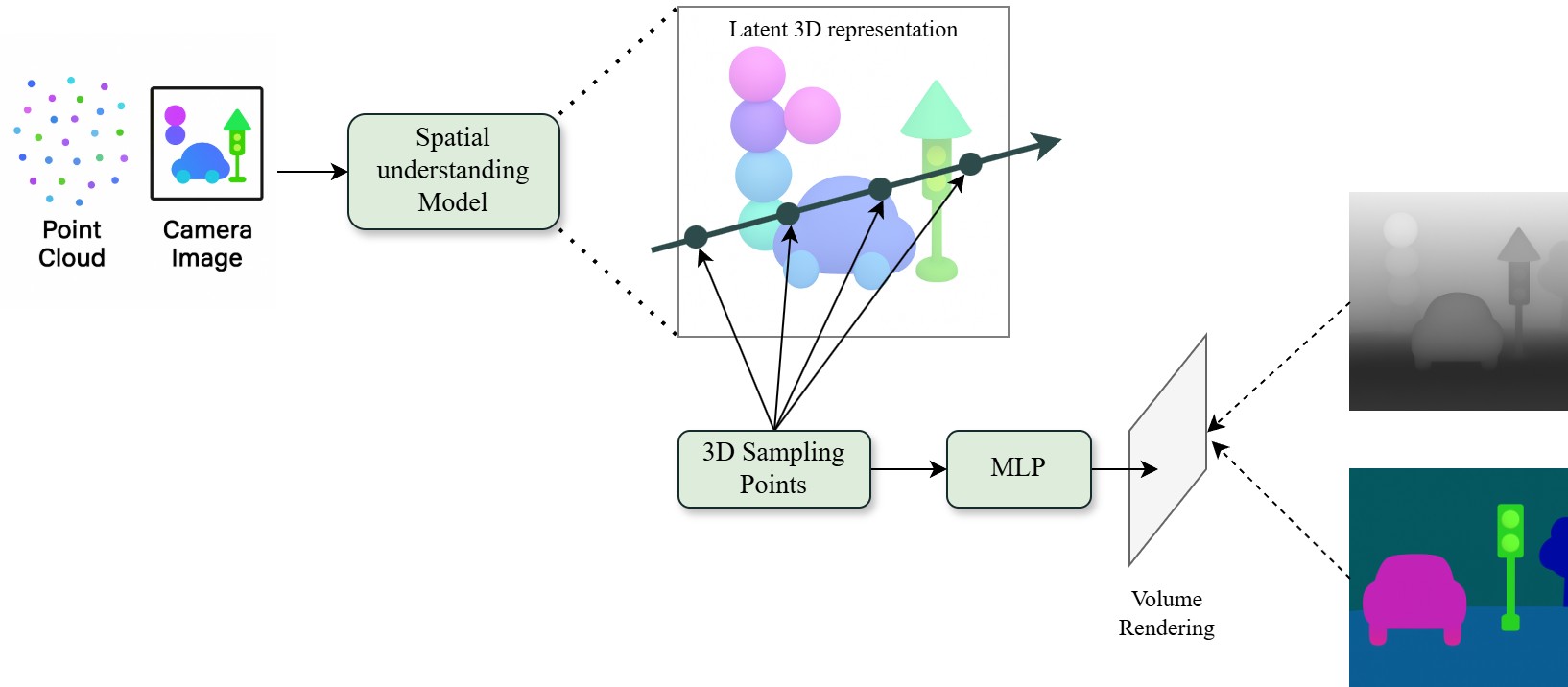}
    \caption{Diagram showing the process of learning 3D representations via 2D supervision using neural rendering. The model generates a latent 3D representation, which is decoded using a NeRF-style rendering technique to produce an image. Supervision is applied in the image space using a depth map and semantic segmentation, guiding the learning process to align with real-world environmental structures.}
    \label{fig:2d_supervision}
\end{figure*}

To mitigate the significant challenge of acquiring densely annotated 3D datasets, numerous models have adopted training paradigms that leverage supervision from more readily available 2D data. Neural Radiance Fields (NeRF) \cite{mildenhall2021nerf} and 3D Gaussian Splatting (3DGS) \cite{kerbl20233d} are two prominent techniques enabling the reconstruction of volumetric scene properties primarily from image-based inputs.

Neural Radiance Fields (NeRF) provide a continuous, implicit volumetric representation by learning a mapping from spatial coordinates (x, y, z) and viewing directions to radiance (color) and density \cite{mildenhall2021nerf}. In the context of autonomous driving, NeRF-inspired methods adapt this formulation for detailed 3D scene understanding. HybridOcc, for example, integrates a transformer-based feature lifting mechanism with NeRF-style depth-guided volume rendering to improve occupancy estimation through ray-based reasoning \cite{zhao2024hybridocc}. Similarly, RenderOcc utilizes volume rendering to project 3D scenes into 2D views, enabling supervision using 2D semantic and depth labels \cite{pan2024renderocc}. S-NeRF++ extends NeRF principles to large-scale simulations, incorporating LiDAR-based refinement to generate diverse and realistic urban scenarios for training perception models \cite{chen2025s}. Furthermore, SelfOcc employs a self-supervised framework, inferring 3D occupancy from temporal image sequences by modeling volumetric predictions as signed distance fields and optimizing them via differentiable multi-view rendering and stereo-guided depth proposals \cite{huang2024selfocc}. The methods discussed utilize NeRF to learn a 3D representation based on 2D labels, as illustrated in \autoref{fig:2d_supervision} which presents their high-level architecture.

The NeRF-style image-space supervision paradigm has also facilitated the integration of large-scale foundational vision models, such as CLIP \cite{radford2021learning} and SAM \cite{kirillov2023segment}, into volumetric frameworks. Techniques like OccNeRF utilize depth maps and SAM-derived segmentation masks to infuse fine-grained semantic knowledge into the learned scene structure, thereby enhancing occupancy prediction capabilities, particularly in LiDAR-free scenarios \cite{zhang2023occnerf}. This cross-modal distillation allows volumetric models to leverage the rich visual and linguistic priors embedded in pre-trained 2D models.

3D Gaussian Splatting (3DGS) offers an alternative, explicit representation by modeling scenes as collections of sparse 3D Gaussian ellipsoids \cite{kerbl20233d}. Each Gaussian encodes spatial position, shape, color, and potentially semantic properties, supporting efficient, high-fidelity rendering and enabling real-time scene manipulation—capabilities highly relevant to autonomous driving systems. RenderWorld utilizes 3DGS to generate self-supervised 3D labels from vision inputs alone, incorporating features like disentangled air/matter encoding and 4D occupancy forecasting to create a unified world model for perception and planning \cite{yan2024renderworld}. GaussianFlowOcc combines temporal flow modeling with a Gaussian Transformer architecture to predict occupancy dynamics, reducing reliance on computationally intensive dense 3D convolutions \cite{boeder2025gaussianflowocc}. Street Gaussians represent dynamic scenes by separating foreground and background Gaussian sets, facilitating object tracking alongside real-time rendering \cite{yan2024street}. Building upon this, GaussianFormer transforms 2D image features into 3D Gaussians using sparse convolutions and attention mechanisms, subsequently aggregating them into voxel-level occupancy predictions \cite{huang2024gaussianformer}. 

\subsubsection{3D Learning via LiDAR-Derived Supervision}
While 2D supervision offers scalability, direct 3D supervision from sensors like LiDAR remains valuable for capturing precise geometric information. Volume rendering techniques allow models to learn from various 3D-related signals, including image segmentation projected onto point clouds, depth maps, and "point painting" methods. Advanced approaches, such as that employed by SurroundOcc , aggregate LiDAR sweeps over time and apply sophisticated reconstruction techniques to generate denser, more complete pseudo-ground truth labels \cite{wei2023surroundocc}. These enhanced labels can then be used to fine-tune volumetric models, integrating richer 3D spatial cues during training.

A distinct approach is exemplified by the UnO (Unsupervised Occupancy) framework, which learns continuous 4D (spatio-temporal) occupancy fields directly from unlabeled LiDAR sequences \cite{agro2024uno}. UnO leverages the physical principles of LiDAR sensing: each returning ray indicates an occupied point, while the path traversed by the ray up to that point represents free space. By aggregating this information across multiple rays and time steps, UnO constructs a self-supervisory signal that guides occupancy learning without explicit labels, enabling generalization to unobserved areas and prediction of occluded regions.

  \subsection{\MakeUppercase{3D Masked Auto Encoders}}

\begin{figure}[htbp]
    \centering
    \includegraphics[width=\columnwidth]{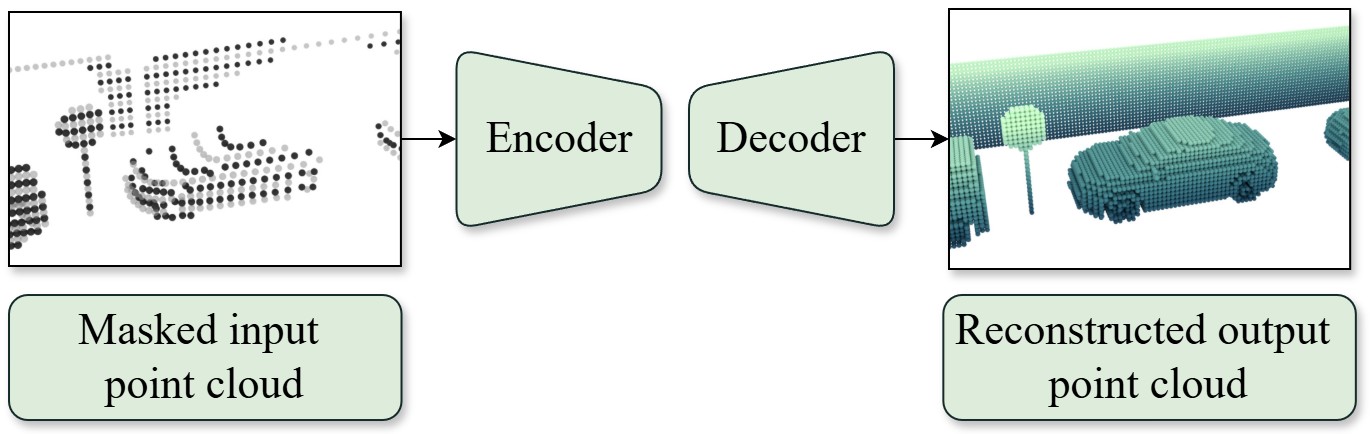}
    \caption{Image from MAELi illustrates the use of 3D Masked Autoencoders to create spatial understanding. A masked point cloud is used as input to a neural network, where an encoder-decoder architecture reconstructs the point cloud \cite{krispel2024maeli}.  }
    \label{fig:maeli}
\end{figure}
Originally successful in natural language processing (e.g., BERT \cite{devlin2019bert}) and 2D vision tasks \cite{he2022masked}, MAEs are now being extended to 3D domains—such as LiDAR and point clouds—where they facilitate the learning of scene-level structure, object boundaries, and spatial relationships crucial for autonomous driving.

For LiDAR point clouds (sparse, irregular, 3D data), masked autoencoding requires different techniques since the data structure is not a regular grid of pixels. One approach is Voxel-MAE \cite{hess2023masked}, which voxelizes the point cloud into a 3D grid and masks a subset of these voxels, then trains the model to reconstruct the missing voxel features or occupancy. By doing so, the model learns to encode the 3D shape such that it can guess points that were removed. Another work, BEV-MAE \cite{lin2024bev}, projects points to the bird’s-eye-view (BEV) plane and masks regions in that BEV space to train a 3D encoder. It adds a point density prediction task to encourage spatial awareness.

Researchers have also proposed enriching the reconstruction targets for point clouds beyond raw points. For example, GeoMAE \cite{tian2023geomae} argues that simply reconstructing coordinates may not fully capture useful geometry; instead, GeoMAE masks groups of points and predicts higher-level \textit{geometric attributes} of the region – such as the centroid, surface normal, or curvature of the local surface, as well as occupancy. These richer prediction targets create a more challenging self-supervised task, which in turn forces the model to learn more discriminative features. \autoref{fig:maeli} illustrates an alternative approach, MAELi  \cite{krispel2024maeli}, which leverages the specific nature of LiDAR data by masking in the sensor’s native spherical coordinate space (taking into account how LiDAR sampling works) and explicitly distinguishing between empty space and occluded regions in its reconstruction loss. By training on large-scale unlabeled LiDAR in this way (even single frames at a time), MAELi taught the model a strong understanding of 3D structure and free-space, providing an excellent initialization for object detection and segmentation.

\subsection{\MakeUppercase{Key Challenges in spatial understanding methods}}

Despite the promise of spatial understanding in enabling unified and fine-grained 3D scene representations, several practical challenges limit its deployment in real-time autonomous driving systems. These challenges span three key areas: computational efficiency, system integration, and supervision. 

\textbf{Latency and Computational Constraints:} Dense volumetric methods, such as semantic occupancy grids, and neural rendering techniques impose substantial latency and memory demands that conflict with the real-time constraints of autonomous driving. Occupancy networks rely on high-resolution voxel grids and costly 3D convolutions, while neural radiance field–based models require dense ray sampling and per-point inference. Although 3D Gaussian Splatting offers faster rendering through explicit primitives, it remains limited in handling dynamic scenes and online updates. These methods, while effective in offline settings, must be aggressively optimized or approximated to meet the latency and resource budgets of embedded perception systems. 

\textbf{Integration Complexity:} Spatial methods often yield representations—such as occupancy maps or radiance fields—not readily aligned with the object-centric abstractions used in planning and tracking. Bridging this gap requires additional processing, such as instance segmentation or volumetric-to-object conversion, which can introduce delays and inconsistencies. 3D Masked Autoencoders (MAEs), while improving internal feature representations and reducing label dependence during training, do not address this core issue. They enhance the quality of downstream perception models but do not fundamentally alter the structure of the spatial output or improve its compatibility with planning modules. As such, the representational mismatch between perception outputs and decision-making inputs persists. Hybrid designs that combine dense spatial context with sparse object-level cues offer a practical compromise, but maintaining consistency and clarity across these modalities remains a significant design challenge. 
 
\textbf{Supervision and 3D ground truth for validation:} Despite the innovations in training methodologies using 2D supervision and LiDAR-based self-supervision, high-quality 3D ground truth remains indispensable for the rigorous evaluation and benchmarking of volumetric perception systems. Large-scale datasets like OpenOccupancy \cite{wang2023openoccupancy} and Occ3D \cite{tian2023occ3d} rely on extensive, often manual, human post-processing and complex reconstruction pipelines to generate accurate voxel-level annotations. This process is inherently time-consuming and costly, underscoring the fact that while alternative supervision strategies can significantly reduce the burden of 3D data collection for \textit{training}, reliable 3D annotations are still critical for assessing model performance and driving progress in the field. 

\begin{table*}[t]
  \centering
  \caption{Summary of Spatial Understanding Methods}
  \label{spatial-methods}
  \begin{tabular}{>{\centering\arraybackslash}p{0.15\linewidth}|>{\centering\arraybackslash}p{0.15\linewidth}
      |>{\centering\arraybackslash}p{0.1\linewidth}
      |>{\centering\arraybackslash}p{0.2\linewidth}
      |>{\centering\arraybackslash}p{0.25\linewidth}
    }
    \hline
    Strategy & Method & Sensors / Inputs & 3D Representation & Supervision Signal \\
    \hline
    Explicit voxel / occupancy& TPVFormer  \cite{huang2023tri}& camera             & Tri-perspective planes $\rightarrow$ 3D voxel grid     & Sparse LiDAR-derived semantic occupancy labels \\
    \cline{2-5}
        & VoxFormer  \cite{li2023voxformer}& camera             & Sparse voxel proposals $\rightarrow$ refined grid      & 2D depth in camera view + 3D semantic occupancy labels \\
    \cline{2-5}
        & OccFormer \cite{zhang2023occformer}& camera             & Dense semantic voxels                                 & 3D semantic occupancy labels \\
    \cline{2-5}
        & SparseOcc \cite{liu2024fully}& camera             & Sparse voxel tokens                                   & 3D semantic occupancy labels \\
    \hline
    NeRF-style neural rendering& HybridOcc \cite{zhao2024hybridocc}& camera             & Semantic density field                                & Multi-camera 2D depth + segmentation (perspective) \\
    \cline{2-5}
        & RenderOcc \cite{pan2024renderocc}& camera             & Semantic density field                                & Multi-camera 2D depth + segmentation (perspective) \\
    \cline{2-5}
        & S-NeRF++  \cite{chen2025s}& camera + LiDAR     & Radiance and density fields                           & Multi-frame photometric consistency + LiDAR depth \\
    \cline{2-5}
        & SelfOcc  \cite{huang2024selfocc}& camera             & Signed-distance field (voxel-wise)                    & Self-supervised multi-view photometric loss \\
    \cline{2-5}
        & OccNeRF \cite{zhang2023occnerf}& camera             & Semantic density field                                & Self-supervised 2D depth, photometric consistency, SAM-derived masks \\
    \hline
    Gaussian-splat world models& RenderWorld \cite{yan2024renderworld}& camera         & 3D Gaussians                                         & Multi-camera 2D depth + segmentation (perspective) \\
    \cline{2-5}
        & GaussianFlowOcc \cite{boeder2025gaussianflowocc}& camera         & 3D Gaussians                                         & Multi-camera, multi-frame 2D depth + segmentation \\
    \cline{2-5}
        & Street Gaussians  \cite{yan2024street}& camera + LiDAR & Fore-/background Gaussians                           & Multi-frame photometric consistency, 2D segmentation \\
    \cline{2-5}
        & GaussianFormer  \cite{huang2024gaussianformer}& camera         & 3D Gaussians                                         & Multi-camera 2D depth + segmentation, 3D semantic occupancy labels \\
    \hline
    LiDAR-driven occupancy& SurroundOcc  \cite{wei2023surroundocc}& camera           & Dense multi-scale semantic voxels                     & Densified LiDAR sweeps + existing 3D detection labels \\
    \cline{2-5}
        & UnO  \cite{agro2024uno}& LiDAR            & Continuous 4D occupancy field                         & Multi-frame unsupervised LiDAR ray-tracing labels \\
    \hline
    3D Masked Auto-Encoders& Voxel-MAE  \cite{hess2023masked}& LiDAR            & Voxel tokens                                         & Reconstruct masked voxel occupancy / point cloud \\
    \cline{2-5}
        & BEV-MAE \cite{lin2024bev}& LiDAR            & BEV feature map                                      & Masked BEV patch reconstruction \\
    \cline{2-5}
        & GeoMAE  \cite{tian2023geomae}& LiDAR            & Voxel tokens                                         & Reconstruct masked point coordinates / normals \\
    \cline{2-5}
        & MAELi  \cite{krispel2024maeli}& LiDAR            & Point cloud                                          & LiDAR ray-tracing freespace/occupancy labels \\
    \hline
  \end{tabular}
  \label{tab:my_double_column_table}
\end{table*}

\section{\MakeUppercase{Multi-sensor Robustness }}

 In the pursuit of reliable autonomous driving, perception systems must operate effectively across a myriad of environmental conditions and scenarios. Multi-sensor robustness refers to the system’s ability to maintain perception accuracy and stability despite environmental variability, sensor noise, or hardware degradation. Different sensors exhibit varying strengths and weaknesses under specific environmental conditions. For instance, cameras may suffer from low visibility in foggy or snowy weather, leading to blurry or occluded views; LiDAR may generate sparse or noisy point clouds under such adverse conditions; while radar typically retains more reliable detection capabilities, though at lower spatial resolution. By integrating complementary sensors such as cameras, LiDAR, and radar, autonomous vehicles can achieve redundancy and cross-validation across data sources, mitigating the limitations of any individual sensor. This multimodal integration is critical to ensure perception reliability in diverse scenarios, as illustrated by the performance trade-offs observed across different modalities under different conditions.

For foundation models in autonomous driving, multi-sensor robustness is especially critical. These models must generalize across domains, weather conditions, and varying sensor setups. Leveraging multimodal data enables the learning of shared representations that are semantically meaningful and geometrically grounded. Such capabilities are vital for improving reliability in edge cases and ensuring scalable deployment across heterogeneous vehicle platforms. 

\subsection{\MakeUppercase{Cross-modality Contrastive Learning}}
   
\begin{figure}[htbp]
    \centering
    \includegraphics[width=\columnwidth]{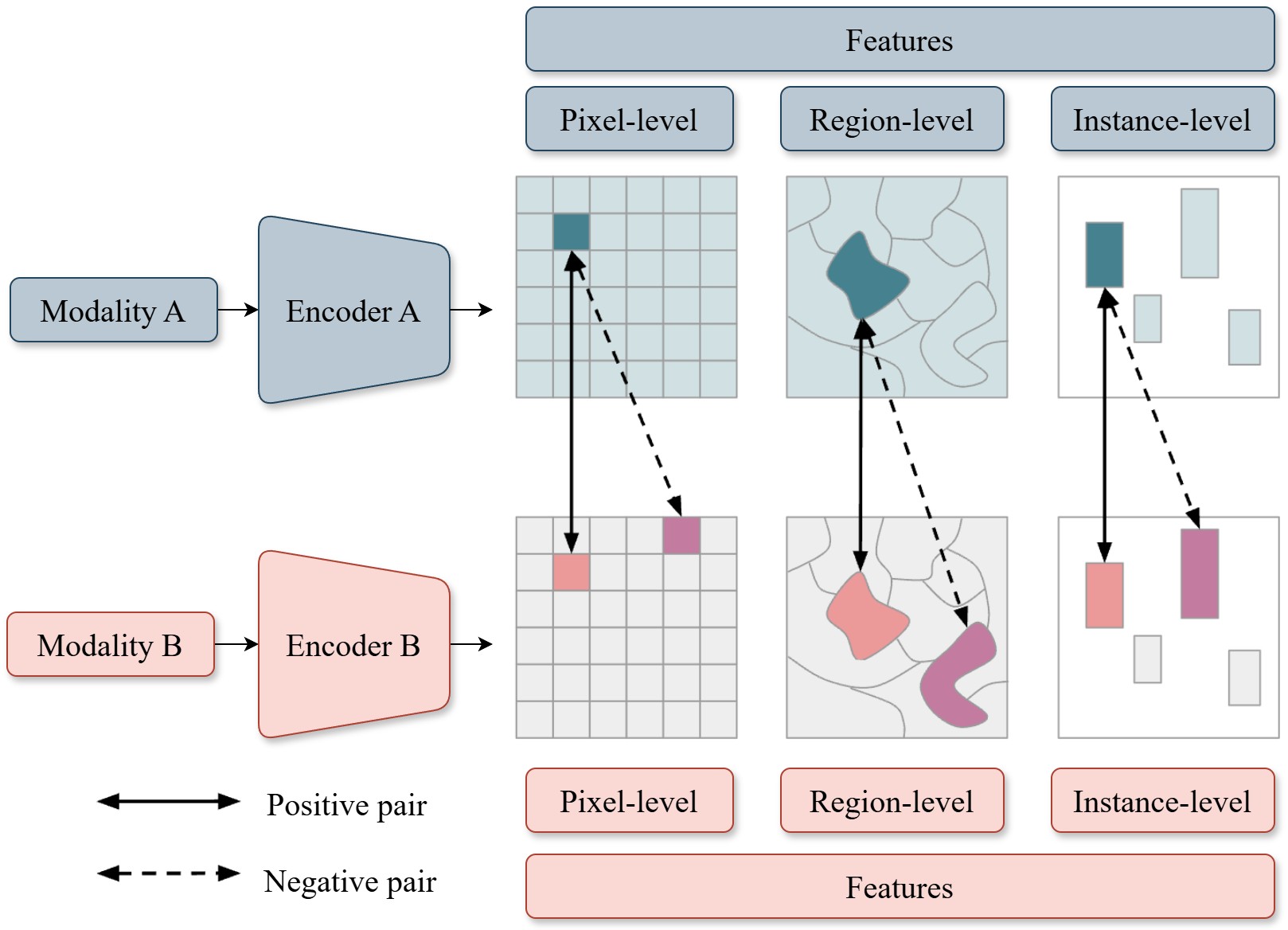}
    \caption{Illustration of cross-modality contrastive learning. Two modalities generate independent features that can be used to construct contrastive pairs, which can be conducted at different granularity levels.}
    \label{fig:multi_modal_cl}
\end{figure}
Cross-modality contrastive learning in autonomous driving aims to learn a unified representation space that integrates complementary information across different sensor types, such as cameras, LiDAR, and radar. By aligning semantically or spatially corresponding features across these diverse sensor inputs, the model is encouraged to develop modality-invariant representations that are robust to environmental changes and sensor limitations. This method utilizes a contrastive loss function such as InfoNCE (Information Noise-Contrastive Estimation) \cite{oord2018representation} to encourage the network to draw together feature representations corresponding to the same spatial or semantic elements across different modalities while pushing apart representations that do not correspond. The InfoNCE loss is defined by the following mathematical formula: 
\[
L_{\mathrm{InfoNCE}} = -\mathbb{E} \left[ 
    \log \frac{ \exp(q \cdot k^{+} / \tau) }
              { \exp(q \cdot k^{+} / \tau) + \sum\limits_{i=1}^{N-1} \exp(q \cdot k_i^{-} / \tau) }
\right]
\]
The components of this equation are defined as follows: 
\begin{itemize}
    \item q (query): The feature representation of an anchor data point from one modality (e.g., a camera image patch).
    \item \textbf{\(k^+\) }(positive key): The feature representation of a data point from another modality that corresponds to the query (e.g., the corresponding LiDAR point cloud).
    \item \textbf{\(k^-\) }(negative keys): The feature representations of non-corresponding data points from the other modality.
    \item \textbf{\(\tau\) }(temperature): A hyperparameter that scales the dot products and controls the sharpness of the distribution. A lower temperature increases the penalty for hard-to-distinguish negative samples. 
\end{itemize}
By systematically constructing positive pairs from co-occurring sensor data that depict the same information and negative pairs from non-corresponding data, the approach compels the model to capture complementary information inherent in each sensor. This process effectively teaches the model to pull the representations of queries and their positive keys closer together in the embedding space while pushing them far apart from the negative keys.  

The specific elements used to construct contrastive pairs can vary across different implementations. \cite{xing2023cross} establishes contrastive pairs by directly associating individual LiDAR points with their corresponding camera image pixels using depth projection. This fine-grained alignment enables pixel-level supervision across modalities but requires accurate calibration and synchronization. In contrast, SuperFlow \cite{xu20244d} and SuperFlow++ \cite{xu2025superflow++} first performs semantic segmentation to group pixels and points into superpixels and supervoxels, respectively. It then conducts contrastive learning at the semantic region level, allowing for more robust and noise-tolerant feature alignment. Meanwhile, ContrastAlign \cite{song2024contrastalign} applies contrastive learning at the instance level by using separate detection heads for LiDAR and camera to extract object-level features. The model aligns these instance features based on detection outputs, which enables object-aware fusion but depends heavily on detection accuracy. As illustrated in \autoref{fig:multi_modal_cl}, these approaches differ in the granularity of the contrastive elements—ranging from pixel-level to region-level to instance-level—highlighting the flexibility of contrastive learning in accommodating various forms of sensor alignment.

\subsection{\MakeUppercase{Cross-modality Knowledge Distillation}}
   
\begin{figure}[htbp]
    \centering
    \includegraphics[width=\columnwidth]{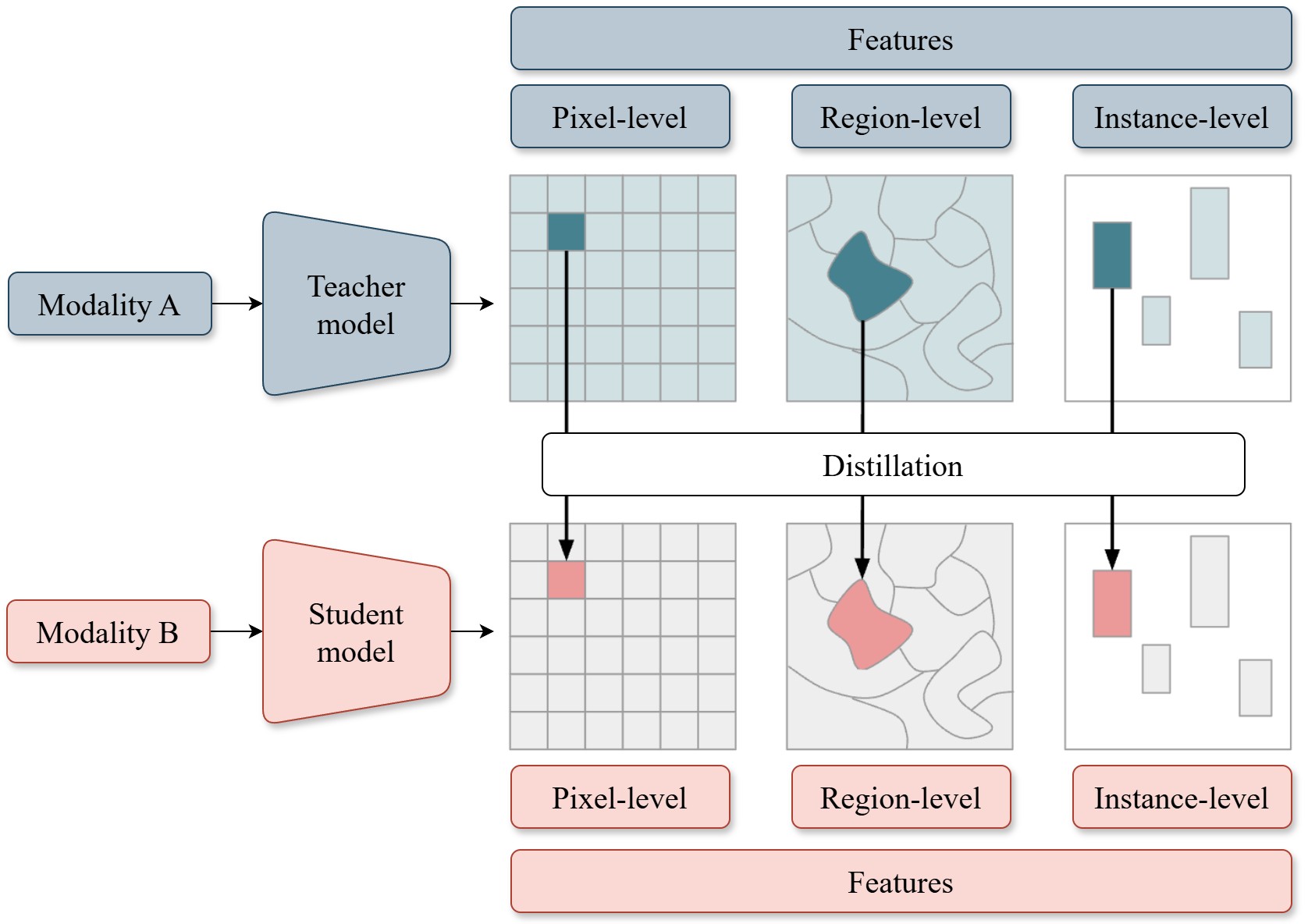}
    \caption{Illustration of cross-modality knowledge distillation. A teacher model using one modality guides the learning process of a student model using another modality, which can be conducted at different granularity levels.}
    \label{fig:multi_modal_distill}
\end{figure}
Cross-modal knowledge distillation uses a high-performing model from one modality to guide a model in another, less informative modality. For example, a LiDAR-based system, with its precise spatial and geometric sensing, can teach a camera-only model to infer spatial cues hard to obtain from images alone. This transfer boosts the camera model’s accuracy and robustness in complex driving scenarios and supports more cost-effective sensor setups without sacrificing perception performance.

Two prevalent approaches to cross-modal knowledge distillation are dense feature distillation and sparse instance distillation \cite{chen2022bevdistill}. Dense feature distillation aims to transfer rich intermediate representations from the teacher to the student model. However, due to the sparsity and modality gap between LiDAR and image features, naive pixel-wise alignment is often suboptimal. To overcome this, DistillBEV \cite{wang2023distillbev} decomposes the BEV feature space into spatial regions and learns region-aware importance weighting, enabling the student to focus on structurally informative areas. Similarly, GAPretrain \cite{huang2023geometric} introduces a LiDAR-guided masking strategy that emphasizes foreground regions and suppresses less-informative background signals during distillation. 

On the other hand, sparse instance distillation targets object-level features. In contrast to dense feature distillation, which transfers pixel- or region-level intermediate features, sparse instance distillation emphasizes the transfer of high-level semantic knowledge by focusing on object instances. BEVDistill \cite{chen2022bevdistill} observes that naive feature-level distillation fails to adequately capture modality differences and often leads to degraded performance. To address this, it proposes using a contrastive InfoNCE loss to encourage better separation of positive and negative instance embeddings across modalities, making it easier for the student to align high-level semantics despite sensing differences. Consequently, methods such as BEVDistill \cite{chen2022bevdistill} and UniDistill \cite{zhou2023unidistill} apply both dense feature and sparse instance distillation, thereby exploiting their complementary strengths and mitigating their respective limitations. \autoref{fig:multi_modal_distill} illustrates the process of cross-modality knowledge distillation applied at different granularity levels.

\subsection{\MakeUppercase{Multi-view Image Consistency}}
\begin{figure}[htbp]
    \centering
    \includegraphics[width=\columnwidth]{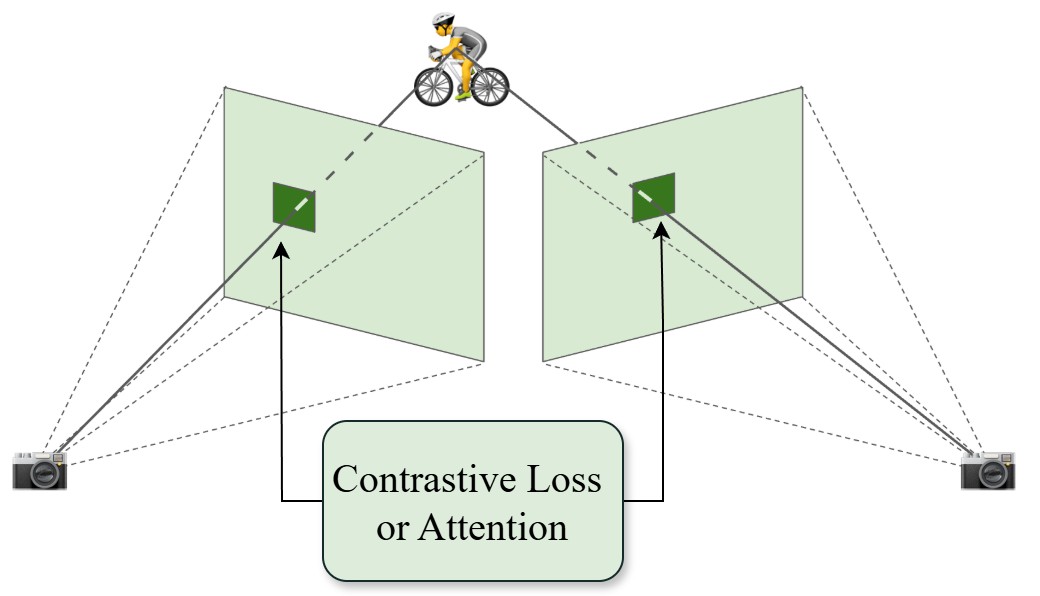}
    \caption{Illustration of multi-view image consistency. Consistency of image features from different views are enforced, or attention between them are applied.}
    \label{fig:multi_view_consistency}
\end{figure}

Multi-camera setups in autonomous driving capture overlapping views of the environment. These overlapping fields of view offer natural supervision signals, as the same scene content is observed from different perspectives. By aligning features from multiple camera views of the same 3D scene as shown in \autoref{fig:multi_view_consistency}, the model can learn viewpoint-invariant representations that are less sensitive to occlusion, camera pose variation, or geometric distortions. For example, Zhang et al. \cite{zhang2022revisiting} address domain generalization in stereo matching by enforcing consistency directly at the feature level. Their approach encourages the learning of domain-invariant representations by explicitly constraining corresponding features in the left and right stereo views to be similar, thus improving robustness across different data domains (e.g., synthetic vs. real). In contrast, WeakMono3D \cite{tao2023weakly} enforces consistency at the prediction level. It transforms the 3D bounding boxes predicted from one camera view into the coordinate system of another using a relative pose matrix and then minimizing the discrepancy between the transformed boxes and the independently predicted boxes from that view with an L1 loss.

Directly contrasting features or predictions across views is simple but sensitive to viewpoint changes and occlusions, as objects may look different or be hidden. Recent methods address this by using attention mechanisms to focus on informative, visible regions, improving robustness to such discrepancies. BEVFormer \cite{li2024bevformer} uses spatial cross-attention where learnable BEV (Bird's-Eye-View) queries attend to features extracted from the multi-camera images. This allows the model to dynamically aggregate relevant information from different views into a unified BEV representation, implicitly handling viewpoint variations and occlusions by focusing attention on visible and informative features across the camera perspectives. EGA-Depth \cite{shi2023ega} introduces a cross-view global attention (EGA) module that allows pixels in one view to attend to informative pixels in another view. This is achieved through an attention mechanism that computes correlations between spatial locations across images, enabling the model to integrate global contextual information from different perspectives. These attention-based mechanisms represent a shift toward more structured and adaptive fusion strategies that exploit cross-view relationships to enhance spatial understanding.

\subsection{\MakeUppercase{Multi-Modal Masked Auto Encoders}} 

\begin{figure}[htbp]
    \centering
    \includegraphics[width=\columnwidth]{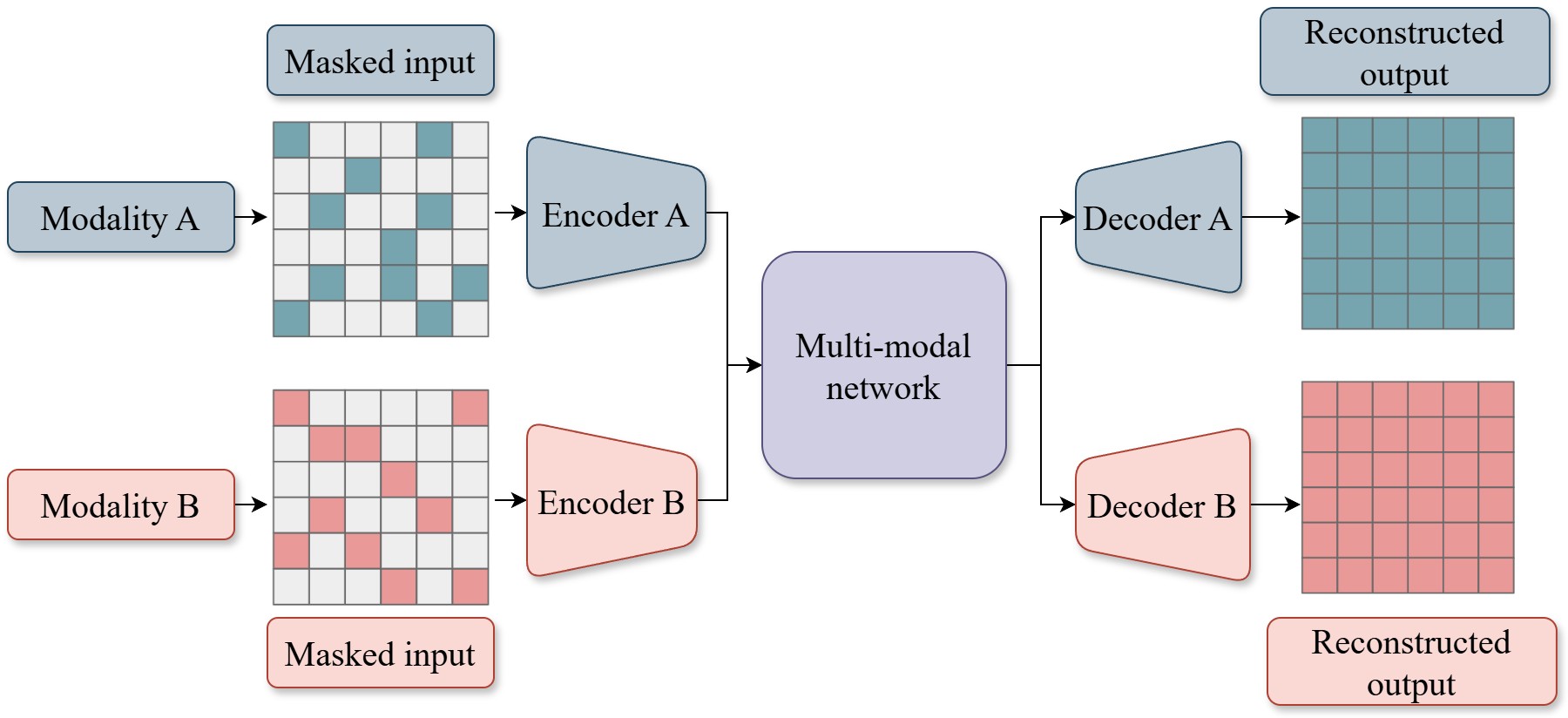}
    \caption{Illustration of multi-modal masked auto encoders. Multi-modal network learns to utilize the information from one modality to reconstruct the other.}
    \label{fig:multi_modal_mae}
\end{figure}
 While Section IV B has discussed masked autoencoders applied to point clouds, another important line of research extends MAEs to handle multi-modal inputs. Recent multi-modal MAE frameworks aim to jointly process heterogeneous data sources such as images, depth maps, and segmentation masks, enabling the model to capture complementary cues and develop stronger, more resilient spatial representations through self-supervised reconstruction, as illustrated in \autoref{fig:multi_modal_mae}. 

MultiMAE \cite{bachmann2022multimae} extends masked autoencoders to jointly learn from multiple modalities—RGB images, depth maps, and segmentation masks. It randomly masks spatial patches across modalities and reconstructs them, fostering cross-modal representations that capture shared structure and semantics in a unified space. Although not trained on LiDAR, its use of depth images—geometrically similar to LiDAR—laid the groundwork for later camera–LiDAR fusion via masked autoencoding.

In terms of camera-LiDAR fusion, PiMAE \cite{chen2023pimae} aligns masked regions across the 3D point cloud and 2D image using geometric projection, which establishes spatial correspondences between 3D points and their associated image pixels. After encoding each modality separately, a shared decoder reconstructs both inputs simultaneously, enforcing that the encoded features are compatible and semantically aligned across 2D and 3D domains. 

Though building upon the idea of joint masked reconstruction, PiMAE still processes point cloud and image data through separate encoder branches, relying on geometric projection for cross-modal interaction. UniM2AE \cite{zou2024unim} achieves a tighter fusion by transforming both camera images and LiDAR point clouds into a shared 3D volumetric representation structured as a bird’s-eye-view grid with vertical height channels. Masking and reconstruction are performed directly within this common space, and a dedicated fusion module is used to facilitate information exchange between the two modalities. These multi-modal MAEs illustrate how self-supervised fusion can create a single foundation model that understands multiple sensors together, potentially simplifying the design of later perception stages.

\subsection{\MakeUppercase{Multi-Modal Diffusion}}  
\begin{figure}[htbp]
    \centering
    \includegraphics[width=\columnwidth]{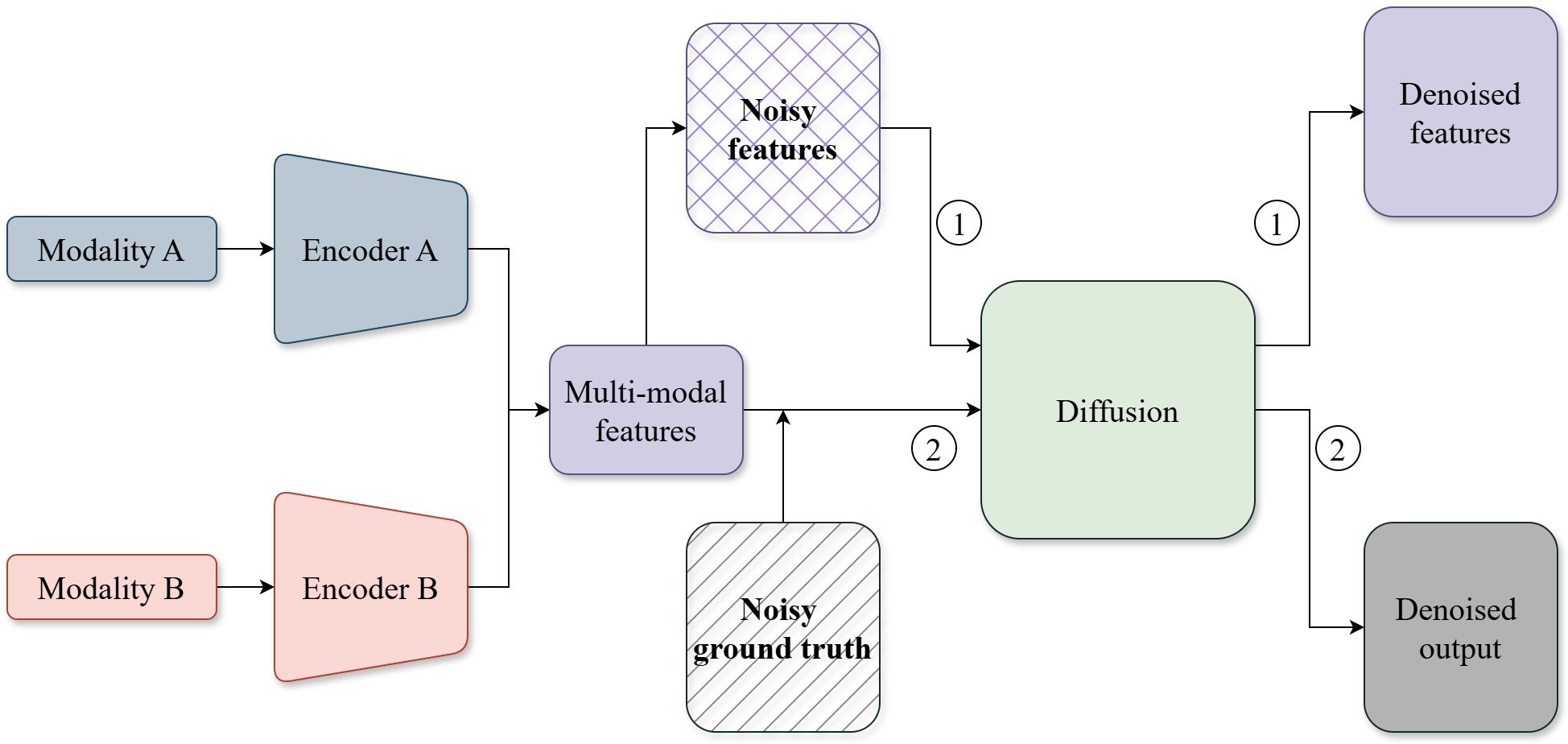}
    \caption{Illustration of multi-modal diffusion. (1) Diffusion is used to denoise augmented multi-modal features. (2) Diffusion is used to denoise augmented ground truth with the assistance of multi-modal features.}
    \label{fig:multi_modal_diffusion}
\end{figure} 
    Diffusion models enhance multi-sensor robustness in autonomous driving by modeling denoising to reduce noise and inconsistencies in multi-modal data. The OccGen framework \cite{wang2024occgen} exemplifies the application of diffusion models at the output level. It adopts a "noise-to-occupancy" generative paradigm, where a 3D Gaussian noise map is progressively refined into a detailed semantic occupancy map. This process is guided by multi-modal features extracted from LiDAR and camera inputs, enabling the model to produce fine-grained occupancy predictions through iterative denoising steps. Such an approach allows for coarse-to-fine refinement, enhancing the granularity and accuracy of the occupancy maps generated. 
   
   DifFUSER \cite{le2024diffusion} advances multi-sensor defusion by emphasizing feature-level denoising. Its conditional diffusion model operates in the latent feature space to refine fused features, using a multi-scale network and conditioning module to handle noisy or incomplete data. The refined features boost downstream tasks like 3D object detection and BEV map segmentation, improving performance and resilience in complex driving scenarios.
   
   Both works illustrate the versatility of diffusion models in addressing the challenges of multi-modal perception and improving robustness in autonomous driving. \autoref{fig:multi_modal_diffusion} shows a comparison of the two diffusion approaches: denoising at the feature level and denoising at the output level.

   Table \ref{tab:multi_sensor_strategies} summarizes representative methods for each multi-sensor robustness strategy discussed above.

\begin{table*}[t]
  \centering
  \caption{Representative Strategies for Multi-Sensor Robustness in Autonomous Driving Perception}
  \label{tab:multi_sensor_strategies}
  \begin{tabular}{>{\centering\arraybackslash}p{0.26\linewidth}
                  |>{\centering\arraybackslash}p{0.15\linewidth}|>{\centering\arraybackslash}p{0.12\linewidth}
                  |>{\centering\arraybackslash}p{0.25\linewidth}
                  |>{\centering\arraybackslash}p{0.12\linewidth}}
    \hline
    \textbf{Strategy} & 
    \textbf{Representative Method} & 
    \textbf{Sensors} & 
    \textbf{Interaction Type} & 
    \textbf{Target Task} \\
    \hline
    Cross-Modality Contrastive Learning& Xing et al. \cite{xing2023cross} & Camera $\leftrightarrow$ Lidar & Pixel-level contrast & Segmentation \\
      \cline{2-5}
    & SuperFlow \cite{xu20244d}, SuperFlow++ \cite{xu2025superflow++} & Camera $\leftrightarrow$ Lidar & Region-level contrast & Pretraining \\
    \cline{2-5}
    & ContrastAlign \cite{song2024contrastalign} & Camera $\leftrightarrow$ Lidar & Instance-level contrast & Detection \\
    \hline
    Cross-Modality Knowledge Distillation& DistillBEV \cite{wang2023distillbev} & Lidar $\rightarrow$ Camera & Region-level distillation & Detection \\
      \cline{2-5}
    & GAPretrain \cite{huang2023geometric} & Lidar $\rightarrow$ Camera & Region-level distillation & Pretraining \\
    \cline{2-5}
    & BEVDistill \cite{chen2022bevdistill} & Lidar $\rightarrow$ Camera & Region- and instance-level distillation & Detection \\
    \cline{2-5}
    & UniDistill \cite{zhou2023unidistill} & Lidar $\leftrightarrow$ Camera & Region- and instance-level distillation & Detection \\
    \hline
    Multi-View Image Consistency& Zhang et al. \cite{zhang2022revisiting} & Camera & Feature-level contrast & Stereo matching \\
      \cline{2-5}
    & WeakMono3D \cite{tao2023weakly} & Camera & Instance-level contrast & Detection \\
    \cline{2-5}
    & BEVFormer \cite{li2024bevformer} & Camera & Feature-level attention & Detection, segmentation \\
    \cline{2-5}
    & EGA-Depth \cite{shi2023ega} & Camera & Feature-level attention & Depth estimation \\
    \hline
    Multi-Modal Masked Autoencoders& PiMAE \cite{chen2023pimae} & Camera, Lidar & Input reconstruction & Pretraining \\
      \cline{2-5}
    & UniM2AE \cite{zou2024unim} & Camera, Lidar & Shared 3D volume feature reconstruction & Pretraining \\
    \hline
    Multi-Modal Diffusion& OccGen \cite{wang2024occgen} & Camera, Lidar & Ground truth denoising & Occupancy prediction \\
      \cline{2-5}
    & DifFUSER \cite{le2024diffusion} & Camera, Lidar & Feature denoising & Segmentation, detection \\
    \hline
  \end{tabular}
\end{table*}

\subsection{\MakeUppercase{Key Challenges in Multi-sensor Robustness}}
To provide a consolidated overview of the discussed techniques, Table \ref{tab:mr_pro_con} offers a comparative analysis of their respective strengths, limitations, and primary applications. Cross-Modality Contrastive Learning excels at learning fine-grained, modality-invariant features that improve robustness to sensor dropouts, though it can be computationally demanding. In contrast, Cross-Modality Knowledge Distillation provides a pathway to high-performing, cost-effective camera-only systems by transferring geometric knowledge from LiDAR, but the student model's performance is bounded by the teacher's quality and potential biases. For systems relying solely on cameras, Multi-View Image Consistency offers a hardware-cheap method for learning viewpoint-invariant features, yet it remains highly sensitive to synchronization and ego-pose inaccuracies. Multi-Modal Masked Autoencoders present a powerful, label-efficient pretraining strategy that is modular and scalable to new sensors, but this flexibility comes at the cost of significant computational resources for pretraining and requires task-specific fine-tuning. Finally, Multi-Modal Diffusion models offer a robust way to mitigate sensor noise and generate calibrated uncertainty estimates, but their iterative sampling process introduces notable inference latency. The following paragraphs discuss common challenges among all methods of multi-sensor robustness.

\textbf{Sensor Reliability}: A significant challenge in multi-sensor robustness methods is that individual sensors are prone to noise and degradation from factors like bad weather, low light, or equipment malfunctions. Errors from one sensor can spread during the fusion process, corrupting the overall environmental model and causing incorrect perception outputs.

\textbf{Calibration and Synchronization}: Achieving robust fusion is complicated by the difficulty of maintaining perfect spatial calibration and temporal synchronization between different sensors. Real-world issues such as calibration drift, different sensor recording patterns (like a spinning LiDAR versus a frame-based camera), asynchronous data capture, and system delays can cause data streams to become misaligned. These misalignments can worsen the propagation of errors across different modalities, making it crucial for fusion mechanisms to be resilient to such imperfections to ensure safe and reliable operation.

\begin{table*}[t]
  \centering
  \caption{Comparison of Multi-Modal Robustness Strategies}
  \label{tab:mr_pro_con}
  \begin{tabular}{>{\raggedright\arraybackslash}p{0.15\linewidth}|>{\raggedright\arraybackslash}p{0.25\linewidth}|>{\raggedright\arraybackslash}p{0.25\linewidth}|>{\raggedright\arraybackslash}p{0.25\linewidth}}
    \hline
    \textbf{Method} & \textbf{Strength} & \textbf{Limitations} & \textbf{Applications} \\
    \hline
    Cross-Modality Contrastive Learning 
      & Learns fine‑grained, modality‑invariant embeddings; improves robustness when one sensor degrades or drops 
      & Compute-heavy when large negative queues are used; performance sensitive to class imbalance 
      & Pretraining shared encoders; multi-sensor fusion; robustness under sensor dropouts \\
    \hline
    Cross-Modality Knowledge Distillation 
      & Enables camera‑only deployment while retaining LiDAR‑level geometry priors 
      & Student inherits teacher bias; struggles if modality gap is large 
      & Camera-only or limited LiDAR stack \\
    \hline
    Multi-View Image Consistency 
      & Encourages viewpoint‑invariant features; hardware‑cheap (cameras only) 
      & Highly sensitive to ego‑pose error and rolling‑shutter time offset; dynamic objects break cross‑view agreement 
      & Multi-camera BEV, depth, 3D detection \\
    \hline
    Multi-Modal Masked Autoencoders 
      & Label‑efficient pretraining; modular—new modalities can be added easily 
      & Pretraining is compute and memory intensive; needs task‑specific finetuning 
      & Universal multi-sensor foundation encoders \\
    \hline
    Multi-Modal Diffusion 
      & Mitigates noisy/partial sensing; generates calibrated uncertainty via sampling variance 
      & Iterative sampling adds inference latency; hyper‑parameter tuning non‑trivial 
      & Semantic occupancy, robust fused features before detection/segmentation \\
    \hline
  \end{tabular}
\end{table*}

\section{\MakeUppercase{Temporal understanding }}
 
Temporal understanding plays a fundamental role in enabling autonomous vehicles to operate safely in dynamic environments. It involves reasoning about how the scene evolves over time—capturing object motion, tracking occlusions, and predicting future events. Unlike static perception, which interprets the environment at a single point in time, temporal perception leverages sequences of observations to build a coherent understanding of motion patterns and interactions among agents.  Its integration into perception addresses key challenges in autonomous navigation.

Temporal modeling supports robust occlusion reasoning and object permanence \cite{yang2024pysical}. By leveraging past observations and motion cues, the system can infer the presence of objects even when they are temporarily hidden, ensuring a continuous and complete environmental model. Temporal consistency further enhances perception by reducing fluctuations in outputs such as bounding boxes and semantic labels. Stable predictions are necessary for reliable planning and smooth control. Finally, temporal understanding allows systems to model how traffic scenes evolve by estimating motion states and predicting future trajectories of agents. These forecasts are critical for anticipatory behaviors like collision avoidance and lane changes.  Table \ref{temporal-methods} consolidates, different method’s strategy, inputs, temporal target, and supervision signal.

\subsection{\MakeUppercase{Temporally Consistent 4D Prediction Models}}

\begin{figure}[htbp]
    \centering
    \includegraphics[width=\columnwidth]{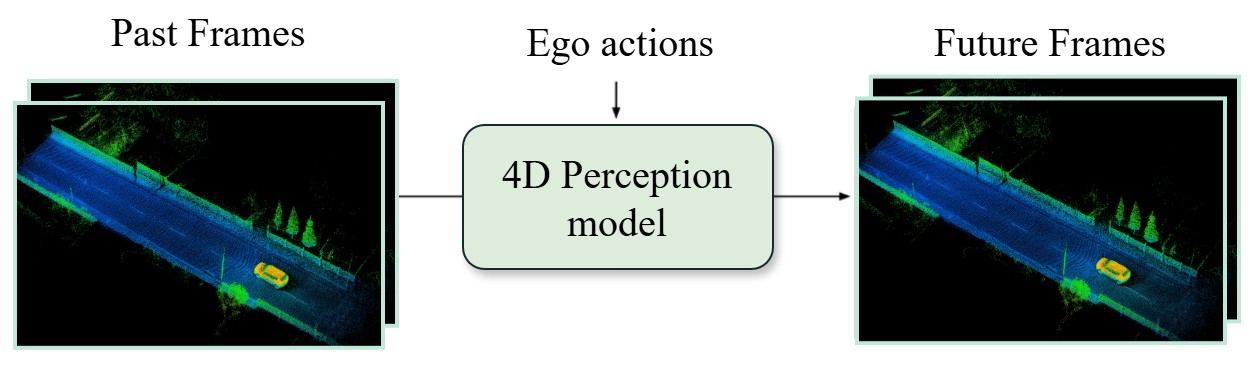}
    \caption{Illustration showing a 4D prediction model that uses past frames and ego actions as inputs to predict future frames, enabling temporally consistent environment forecasting for autonomous driving.}
    \label{fig:4d_temporal}
\end{figure}

Predictive modeling of future scene dynamics is a critical capability for autonomous driving systems. A central challenge lies in forecasting the evolution of complex, dynamic environments from sequences of sensor observations. Recent methodologies increasingly emphasize the principle of temporal consistency, aiming to learn representations that accurately capture how scenes change over time. \autoref{fig:4d_temporal} illustrates the inputs and outputs of a temporally consistent 4D prediction model. 

\subsubsection{Learning Temporal Representations}

Several approaches leverage temporal consistency through self-supervised or supervised learning on sequential sensor data. For instance, UnO \cite{agro2024uno} utilizes sequential LiDAR data to learn continuous 4D occupancy fields. By training the model to forecast future occupancy states and comparing these predictions against subsequent LiDAR measurements, UnO enforces temporal consistency, enabling it to learn spatio-temporal representations encoding both geometry and motion. This inherent temporal modeling underpins its ability to infer dynamic scene structures and generalize to various downstream tasks.

Similarly, Visual Point Cloud Forecasting \cite{yang2024visual} methods focus on predicting future 3D point clouds primarily from sequences of camera images, sometimes augmented with past LiDAR data. These models infer temporal dynamics by processing time-aligned visual features, often employing architectures like temporal convolutions or Transformers to maintain sequential dependencies. The predicted temporal features are then lifted into coherent 3D future representations using techniques such as depth estimation or volumetric rendering. The core temporal aspect involves projecting sequential 2D observations into consistent 3D forecasts.

UniWorld  \cite{min2023uniworld} adopts a pre-training strategy, learning spatio-temporal features by forecasting 4D occupancy from sequences of image-LiDAR pairs. Temporal consistency is enforced through future occupancy prediction objectives, and multi-frame fusion mechanisms are employed to effectively integrate temporal context. This focus on temporal understanding during pre-training enhances performance on dynamic downstream tasks by embedding temporal supervision directly within the learned representations.

\begin{table*}[t]
  \centering
  \caption{Summary of 4D Prediction and Contrastive Learning Methods}
  \label{temporal-methods}
  \begin{tabular}{>{\centering\arraybackslash}p{0.1\linewidth}|>{\centering\arraybackslash}p{0.18\linewidth}
    |>{\centering\arraybackslash}p{0.16\linewidth}
    |>{\centering\arraybackslash}p{0.16\linewidth}
    |>{\centering\arraybackslash}p{0.3\linewidth}}
    \hline
    \textbf{Strategy} & \textbf{Method} & \textbf{Sensors / Inputs} & \textbf{Temporal Target} & \textbf{Learning Objective (Supervision)} \\
    \hline
    4D Prediction Models& UnO (Unsupervised Occupancy Fields)  \cite{agro2024uno}& Sequential LiDAR sweeps
      & Continuous 4‑D occupancy field
      & Self-supervised: Pseudo-labels from LiDAR ray casting (classifying points along the ray as occupied or free) \\
    \cline{2-5}
      & Visual Point‑Cloud Forecasting (ViDAR) \cite{yang2024visual}& Camera video (+ past LiDAR optional)
      & Future 3‑D point clouds
      & Self-supervised: Predicts future LiDAR point clouds from video by jointly supervising semantics, 3D structure, and dynamics \\
    \cline{2-5}
      & UniWorld \cite{min2023uniworld}& Image–LiDAR sequences
      & 4‑D semantic occupancy
      & Self-supervised: Pre-trained through masked prediction of future occupancy and geometry \\
    \hline
    Diffusion Models& CoPilot4D (diffusion world-model) \cite{zhang2023learning}& Past multi-sensor data + planned actions
      & BEV token sequence (stochastic)
      & Trained via discrete diffusion conditioned on the sequence of planned actions \\
    \cline{2-5}
      & BEVWorld \cite{zhang2024bevworld}& Multi-sensor history (camera, LiDAR)
      & Unified BEV latent scene
      & Latent-space diffusion model reconstructs BEV representation from historical sensor data \\
    \cline{2-5}
      & Drive‑WM \cite{wang2024driving}& Multi-view images
      & Multi-view future video
      & Image diffusion conditioned on temporal and camera-view tokens to generate future video frames \\
    \hline
    Contrastive Learning& TARL (Temporal contrastive) \cite{nunes2023temporal}& Sequential 3‑D LiDAR scans
      & Segment-level descriptors
      & Siamese network with contrastive learning. Uses ego-motion to create positive pairs of temporally close LiDAR segments for learning robust representations \\
    \cline{2-5}
      & SuperFlow \cite{xu20244d}& LiDAR + camera sequence
      & Superpoint features \& Scene Flow
      & Contrastive learning on spatial and temporal correspondences between points in LiDAR and camera data \\
    \cline{2-5}
      & SuperFlow++  \cite{xu2025superflow++}& Multi-camera + LiDAR
      & Superpoint features \& Scene Flow
      & Extends SuperFlow by adding cross-view consistency loss to the contrastive learning objective \\
    \cline{2-5}
      & COMPASS \cite{ma2022compass}& RGB Video, IMU, LiDAR
      & Shared cross-modal embedding
      & Graph-based contrastive learning across spatial, temporal, and modal dimensions to learn unified representation \\
    \hline
  \end{tabular}
  \label{tab:4d_prediction_methods}
\end{table*}

\subsubsection{Diffusion Models for Probabilistic Future Prediction}

While the aforementioned methods advance temporal modeling, capturing the inherent uncertainty and multi-modality of future scenarios remains a significant challenge. Driving environments are stochastic; multiple plausible future states can arise from a given history (e.g., a vehicle turning left, right, or proceeding straight). Deterministic prediction models often struggle with this, potentially yielding unrealistic averaged outcomes.

Diffusion probabilistic models offer a compelling alternative due to their capacity to represent complex, multi-modal probability distributions. By learning to reverse a gradual noising process, diffusion models can capture the full data distribution. Initiating the reverse process from different noise samples allows them to generate diverse and plausible future scenarios, which is crucial for robust planning under uncertainty.

CoPilot4D \cite{zhang2023learning} exemplifies the application of diffusion models to multi-modal future forecasting in autonomous driving. It employs a spatio-temporal Transformer integrated within a discrete diffusion framework to predict future sequences of bird's-eye view (BEV) tokens representing the environment's state. Crucially, CoPilot4D conditions its predictions on both past sensor data and planned ego-vehicle actions (e.g., future poses). This action conditioning enables counterfactual reasoning – simulating the likely outcomes of different potential maneuvers. By leveraging tokenization and discrete diffusion on unlabeled point cloud data, CoPilot4D effectively adapts generative modeling techniques from other domains to learn action-conditioned 4D world models, achieving state-of-the-art results in point cloud forecasting.

Other diffusion-based approaches further explore this direction. BEVWorld \cite{zhang2024bevworld} extends diffusion modeling to a unified BEV latent space derived from multi-sensor inputs, using a spatio-temporal Transformer conditioned on action tokens for consistent multi-modal scene forecasting. Drive-WM \cite{wang2024driving} combines image diffusion with temporal and multi-view modeling to forecast future driving scenes from multiple perspectives, facilitating video-based planning through controllable, consistent sequence generation.

\subsection{\MakeUppercase{Temporal contrastive learning}}
\begin{figure}[htbp]
    \centering
    \includegraphics[width=\columnwidth]{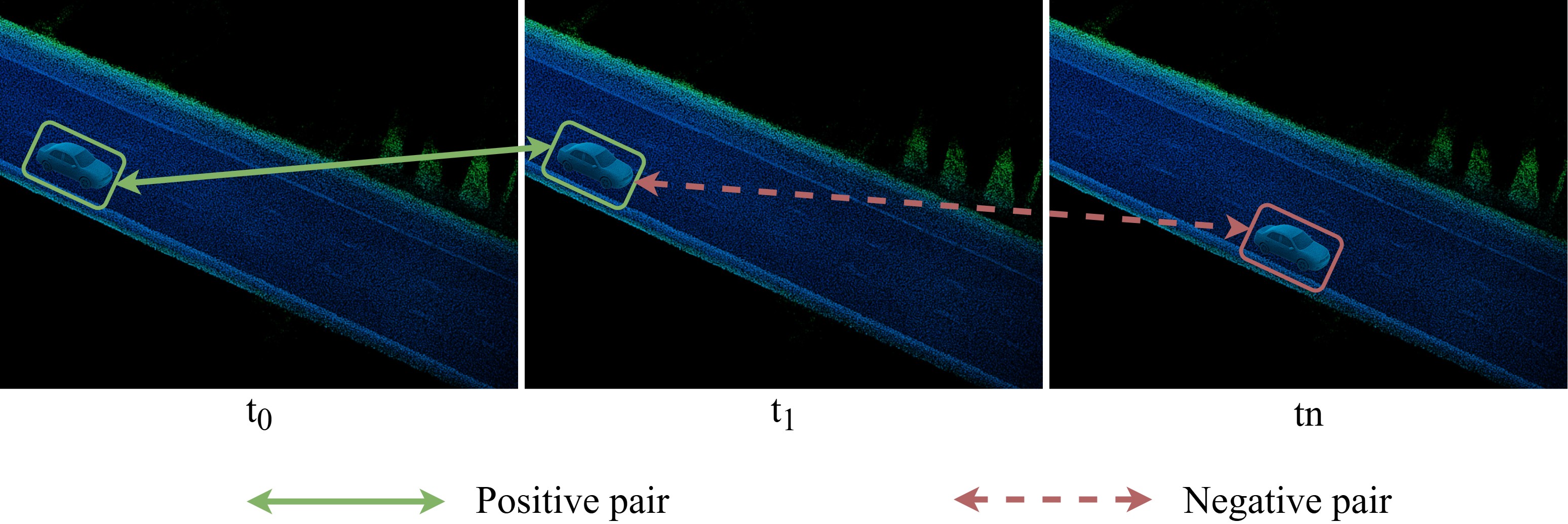}
    \caption{Illustration showing positive pairs (t0, t1) as temporally adjacent frames and negative pairs (t1, tn) as frames with a larger temporal gap, illustrating the core concept of temporal contrastive learning.}
    \label{fig:temporal_constrastive}
\end{figure} 

 In the context of autonomous driving, contrastive learning is increasingly leveraged to learn temporally consistent and motion-aware representations by capitalizing on the sequential nature of sensor streams. As the ego-vehicle navigates through the environment, it continuously observes objects and scenes from varying perspectives due to its own motion or the motion of other agents. These temporal variations act as implicit augmentations, where different views of the same object across time are used as positive pairs. In contrast, observations of different objects or temporally distant views of the same object can serve as negative pairs. \autoref{fig:temporal_constrastive} depicts an example of how positive and negative pairs can be chosen for temporal contrastive learning. 

This methodology transforms temporal coherence in the physical world into a powerful self-supervisory signal. It allows models to learn representations that are robust to viewpoint shifts while preserving sensitivity to the identity and dynamic state of observed entities. Such temporally grounded representations are particularly beneficial in dynamic and safety-critical environments like autonomous driving, where understanding object behavior and consistency over time is crucial.

Several advanced frameworks exemplify this approach. TARL (Temporal Association Representation Learning) \cite{nunes2023temporal} targets 3D LiDAR data and utilizes ego-motion to associate object segments across consecutive scans. It segments potential objects using unsupervised clustering methods and trains a Siamese network to maximize the similarity between point-level features and their aggregated temporal representations. The inclusion of a Transformer-based projection head further enhances the model's capacity to learn intra-object relationships, resulting in temporally aligned features suitable for tasks like segmentation.  SuperFlow \cite{xu20244d} and SuperFlow++ \cite{xu2025superflow++} extend this concept to multimodal inputs, particularly LiDAR-camera pairs. They incorporate a flow-based contrastive learning module that compares superpoint features across time and modality. SuperFlow integrates spatial and temporal contrastive losses to align LiDAR superpoints and image superpixels, while SuperFlow++ introduces multi-camera consistency constraints to enforce robust feature learning across viewpoints. COMPASS (Contrastive Multimodal Pre-training for Autonomous Driving) \cite{ma2022compass} proposes a unified framework for multimodal contrastive learning. It constructs a graph connecting RGB images, depth maps, and optical flow across time steps and maps them into a shared spatial and motion pattern space. Contrastive objectives are applied across spatial, temporal, and spatiotemporal dimensions, enabling rich, context-aware feature representations that capture both scene geometry and dynamic behavior.

Together, these approaches showcase how contrastive learning, when combined with the temporal structure of driving data and multimodal sensor streams, enables the development of robust, temporally coherent perception systems for autonomous vehicles.

\subsection{\MakeUppercase{Key Challenges in Temporal Understanding for Autonomous Driving}}

Temporal understanding in this context is grounded in three core, interdependent challenges: object permanence and occlusion reasoning, motion prediction and trajectory forecasting, and temporal consistency.

\textbf{Object Permanence and Occlusion:} Object permanence and occlusion reasoning address the reality that road users are often temporarily hidden by other objects. In the absence of temporal modeling, perception systems can lose track of these agents, posing a major safety risk. By integrating information from past observations and analyzing motion cues, temporal models can infer the continued existence and trajectory of occluded objects, ensuring a complete and reliable world model for planning and collision avoidance.

\textbf{Uncertainty in Motion Prediction:} Motion prediction and trajectory forecasting are essential for anticipatory driving. Autonomous vehicles must model how agents will move in the future, often in highly uncertain and multi-modal traffic scenarios. Deterministic models struggle to capture the diversity of plausible futures, typically averaging outcomes and producing unrealistic predictions. Probabilistic models, such as diffusion-based approaches, address this by generating a distribution over possible futures, but introduce additional computational and representational complexity, as well as challenges in producing physically plausible and actionable predictions.

\textbf{Ensuring Temporal Consistency:} Temporal consistency is vital for stable operation. Frame-by-frame processing can yield noisy or jittery outputs in object detection, tracking, or segmentation, which can propagate to downstream modules and result in erratic vehicle behavior. Temporal smoothing and consistency constraints ensure that predictions are coherent and physically plausible across time, supporting smooth and trustworthy control.

\textbf{Real-Time Latency Constraints:} A core challenge in temporal 4D perception is realizing recurrent world models that maintain a compact, stable BEV or semantic-occupancy state and update it efficiently under strict on-vehicle latency budgets. The model must capture long-horizon dependencies via state-space recurrence or memory-augmented transformers while bounding context through sliding windows, sparse attention, and selective key-frame or event-driven updates. Robust recurrence further hinges on precise cross-sensor alignment—tight time synchronization, ego-motion compensation, and periodic re-initialization—to limit drift and error accumulation. Meeting these requirements within power and memory constraints demands aggressive efficiency measures, including knowledge distillation, token/voxel pruning, and budgeted mixture-of-experts routing. Complementary systems optimizations—low-precision (INT8/FP8) execution with calibration, operator fusion, cache reuse, and graph capture on accelerators—are likewise essential. Integrating these elements cohesively remains difficult, making recurrent 4D prediction a central bottleneck for practical, real-time deployment. 

In practice, these challenges are compounded by the uncertainty of real-world dynamics, imperfect sensor calibration and synchronization, and the substantial computational demands of high-dimensional spatio-temporal representations. The integration of robust temporal understanding into real-time, safety-critical systems remains an open problem, often requiring complex hybrid system architectures. Addressing these interlinked challenges will require advances in model design, optimization, hardware acceleration, and evaluation methodologies.

\section{\MakeUppercase{Challenges and Future work}}
Despite the significant promise of foundation models for autonomous driving perception, several challenges remain unresolved in the path toward building practical, reliable, and deployable systems. This section highlights the major limitations and directions for future advancement, with a primary emphasis on how the core capabilities can be integrated into a unified framework, while also providing dedicated discussions of each individual capability—generalized knowledge, spatial awareness, multi-sensor robustness, and temporal understanding—to highlight both their unique challenges. The issue of real-time latency is examined in the end, underscoring the deployment challenges posed by the high computational demands of foundation models.

\subsection{\MakeUppercase{Integration of Core Capabilities}}
While promising solutions have emerged to address individual aspects of foundational models, no existing system seamlessly integrates all four capabilities—generalized knowledge, spatial reasoning, multi-sensor robustness, and temporal understanding—into a unified, real-time operational framework. Examples are shown in Table \ref{tab:core-capabilities}. Although research has explored combinations of these capabilities, a holistic, scalable solution remains elusive. SEAL \cite{liu2023segment} serves as a representative example that both distills knowledge from vision foundation models (VFMs) and incorporates temporal and multi-modal contrastive learning. This design enables the utilization of generalized knowledge with enhanced multi-sensor robustness and temporal understanding. However, system-level integration and latency optimization remain insufficiently explored and require further investigation to ensure real-time feasibility and reliability. 
Hybrid system architectures represent another pragmatic interim solution, integrating capability-rich foundation models with traditional, highly optimized real-time autonomous driving pipelines. In such systems, the foundational model might handle less time-critical, higher-level reasoning tasks or operate at a reduced frequency. Simultaneously, conventional components manage continuous, low-latency perception and control functions, thus balancing advanced capabilities with stringent real-time constraints \cite{tian2024drivevlm}. While an engineering compromise, this approach bridges the current gap. However, future work must strive towards developing real-time capable foundation models that can holistically integrate all their advanced perception, reasoning, and fusion capabilities without relying on such dual pipelines, thereby achieving a more unified and inherently robust system.

\begin{table*}[t]
  \centering
  \caption{Coverage of Core Capabilities in Recent Foundation Models for Autonomous Driving
  }
  \label{tab:core-capabilities}
  \begin{tabular}{l|c|c|c|c}
    \hline
    \textbf{Model / Method} & \textbf{Generalized Knowledge} & \textbf{Spatial Understanding} & \textbf{Multi-Sensor Robustness} & \textbf{Temporal Understanding} \\
    \hline
    SLidR \cite{sautier2022image} / SEAL \cite{liu2023segment}& X &   & X &   \\
    OVO \cite{tan2023ovo} / CLIP2Scene \cite{chen2023clip2scene}& X & X &   &   \\
    SAL \cite{ovsep2024better} / SAM4UDASS \cite{10366854}& X & X &   &   \\
    UP-VL \cite{najibi2023unsupervised} / OpenSight  \cite{zhang2024opensight}& X & X &   &   \\
    OccNeRF \cite{zhang2023occnerf}& X & X &   &   \\
    OmniDrive \cite{wang2024omnidrive} / DriveGPT4  \cite{xu2024drivegpt4}& X & X &   & X \\
    EMMA \cite{hwang2024emma}& X & X &   & X \\
    SelfOcc  \cite{huang2024selfocc}&   & X &   & X \\
    RenderWorld \cite{yan2024renderworld}&   & X &   & X \\
    GaussianFlowOcc \cite{boeder2025gaussianflowocc}&   & X &   & X \\
    UnO \cite{agro2024uno}&   & X &   & X \\
    SuperFlow++ \cite{xu2025superflow++}&   &   & X & X \\
    COMPASS \cite{ma2022compass}&   &   & X & X \\
    BEVDistill \cite{chen2022bevdistill} / UniDistill \cite{zhou2023unidistill}    &   &   & X &   \\
    BEVFormer  \cite{li2024bevformer}                   &   & X & X &   \\
    PiMAE \cite{chen2023pimae} / UniM2AE \cite{zou2024unim}&   & X & X &   \\
    OccGen \cite{wang2024occgen} / DifFUSER \cite{le2024diffusion}&   & X & X &   \\
    UniWorld \cite{min2023uniworld}&   & X & X & X \\
    CoPilot4D \cite{zhang2023learning} / BEVWorld \cite{zhang2024bevworld}&   & X & X & X \\
    TARL \cite{nunes2023temporal}&   &   &   & X \\
    \hline
  \end{tabular}
\end{table*}

\subsection{\MakeUppercase{Limitations of Current Benchmarks and the Need for Real-World Evaluation}}
In addition, although numerous benchmarks exist for evaluating autonomous driving systems \cite{sun2020scalability, caesar2020nuscenes, chang2019argoverse, geiger2012we, yu2020bdd100k}, most focus on general-case scenarios and overlook rare or safety-critical corner cases—precisely the scenarios where foundation models must prove their robustness and practical utility. This gap leads to a mismatch between benchmark performance and real-world reliability. As a result, research efforts often prioritize improving average-case metrics rather than addressing the most consequential situations. To foster meaningful progress, better benchmarks must be developed that systematically target these core capabilities, and future work should emphasize tackling such difficult scenarios through deeper investigation and targeted evaluation.

Benchmarks such as KITTI \cite{geiger2012we}, nuScenes \cite{caesar2020nuscenes}, and the Waymo Open Dataset \cite{sun2020scalability} have propelled progress in perception tasks under common driving conditions. However, these datasets primarily capture typical environments—clear weather, daylight, and predictable traffic—and lack adequate representation of rare edge cases such as erratic pedestrian behavior, debris, unusual lighting, or extreme weather. Consequently, models trained and evaluated on these benchmarks may perform well on average-case metrics (e.g., mAP, IoU) but fail under conditions critical to safety in real-world deployment.

While some benchmarks now aim to include rare scenarios, they are typically generated in simulation environments. For example, DeepAccident \cite{Wang_2023_DeepAccident}, AutoScenario \cite{lu2024realistic}, and PeSOTIF \cite{peng2023pesotif} offer stress tests involving rare or safety-critical events using simulators like CARLA \cite{dosovitskiy2017carla}. Although valuable for controlled analysis, these synthetic datasets often lack the fidelity needed to fully replicate complex real-world interactions, such as unmodeled sensor artifacts or unpredictable human behavior. Thus, they are not ideal for final validation. Table \ref{tab:benchmark} provides a summary of benchmarks designed to evaluate these core capabilities across various conditions.

To bridge this gap, future evaluation paradigms must move beyond static datasets to scenario-based testing that incorporates both curated real-world data and synthetic augmentation. This includes perturbing real logs to simulate sensor failure, occlusion, or adverse weather, and embedding metrics that explicitly test robustness under these conditions. Such targeted benchmarks would shift optimization away from aggregate accuracy toward deployment-relevant resilience.

Most importantly, benchmark design should align with the validation strategies used for real-world deployment. Autonomous vehicles must demonstrate robust performance not only in simulated stress tests but across millions of real-world miles to uncover unforeseen hazards and long-tail edge cases. Lessons from these deployments can inform dynamic benchmark curation, creating evolving corner-case libraries that ground evaluation in operational reality. Embedding these insights into benchmarking pipelines will ensure that systems excelling in benchmark settings are also prepared for the full complexity and unpredictability of public roads.

\begin{table*}[t]
  \centering
  \caption{Summary of Benchmarks for Evaluating Capabilities in Autonomous Driving Perception}
  \label{tab:benchmark}
  \begin{tabular}{>{\raggedright\arraybackslash}p{0.12\linewidth}|>{\raggedright\arraybackslash}p{0.18\linewidth}|>{\raggedright\arraybackslash}p{0.1\linewidth}|>{\raggedright\arraybackslash}p{0.3\linewidth}|>{\raggedright\arraybackslash}p{0.2\linewidth}}
    \hline
    \textbf{Benchmark} & \textbf{Capabilities Measured} & \textbf{Inputs/Sensors} & \textbf{Measures / Tasks} & \textbf{Type} \\
    \hline
    PeSOTIF \cite{peng2023pesotif}& Generalized knowledge & cameras & Uncertainty in rare scenarios & Real \\
    \hline
    DeepAccident \cite{Wang_2023_DeepAccident}& Generalized knowledge & cameras, LiDAR & 3D object detection, tracking, BEV segmentation in accident-focused scenarios & Simulation \\
    \hline
    NuPrompt \cite{wu2025language} & Generalized knowledge, Spatial understanding, Temporal understanding & cameras, LiDAR & Prompt-based 3D object tracking & Real \\
    \hline
    AnoVox \cite{bogdoll2024anovox} & Spatial understanding, Temporal understanding & cameras, LiDAR & Temporal anomaly detection and 3D semantic occupancy & Simulation \\
    \hline
    OpenOccupancy \cite{wang2023openoccupancy}& Spatial understanding & cameras, LiDAR & 3D semantic occupancy & Real-world data with synthetic densification \\
    \hline
    Occ3D  \cite{tian2023occ3d}& Spatial understanding & cameras, LiDAR & 3D semantic occupancy & Real-world inputs with automated voxel-labeling \\
    \hline
    MSC-Bench \cite{hao2025msc} & Multi-sensor robustness & cameras, LiDAR & 3D detection with 16 corruption types & Real data with synthetic corruptions \\
    \hline
    KITTI-C, nuScenes-C, Waymo-C \cite{dong2023benchmarking} & Multi-sensor robustness & cameras, LiDAR & 3D object detection under 27 corruption types & Real data with synthetic corruptions \\
    \hline
    Robo3D \cite{kong2023robo3d} & Multi-sensor robustness & LiDAR & 3D object detection and 3D semantic segmentation with 8 corruption types/severity levels & Real data with synthetic corruptions \\
    \hline
    RoboBEV \cite{xie2023robobev} & Multi-sensor robustness & cameras & BEV detection under camera image corruption & Real data with synthetic corruptions \\
    \hline
    ``Unity is Strength?'' \cite{jin2024unity} & Multi-sensor robustness & cameras, LiDAR & 3D object detection under sensor dropout/corruption & Real data with synthetic corruptions \\
    \hline
    R‑U‑MAAD \cite{wiederer2022benchmark} & Temporal understanding & object-level positions & Detection of anomalous multi-agent trajectories & Unsupervised metrics \\
    \hline
  \end{tabular}
\end{table*}

\subsection{\MakeUppercase{Real-Time Latency Mitigation}}
While large foundation models bring strong generalization and reasoning capabilities, their substantial computational overhead makes them difficult to deploy in practice. High inference latency makes these models hard to meet the real-time demands of autonomous driving without optimization. Knowledge distillation, model quantization and parameter pruning are significant approaches commonly used for model compression and acceleration \cite{cheng2017survey}. For knowledge distillation, in addition to the works mentioned in previous sections, many other studies focus on general foundation models \cite{hsieh2023distilling, wu2022tinyvit, li2023distilling} and have not been thoroughly explored in the context of autonomous driving. Similarly, model quantization, which reduces the numerical precision of model parameters and activations, and parameter pruning, which removes redundant or less important weights, have both been extensively studied for general foundation models \cite{dettmers2022gpt3, li2023repq, sun2023simple, yu2023x}. However, their application in autonomous driving remains limited and holds significant potential for further development. More specifically, careful implementation is required to minimize potential accuracy degradation, which is critical in safety-sensitive applications like autonomous driving, and more comprehensive research is necessary to fully understand and optimize their use in this domain.

In addition to algorithmic optimization, complementary advances include the development of specialized hardware accelerators, enhanced memory systems (with higher bandwidth and capacity), and improved interconnects (to facilitate faster data movement). This includes custom Application-Specific Integrated Circuits (ASICs), Field-Programmable Gate Arrays (FPGAs), and novel architectures like Grok's  \cite{moon2024lpu} Language Processing Unit (LPU) designed for predictable low-latency AI inference, or the leveraging of increasingly powerful commercial edge AI platforms. These hardware solutions are tailored to the specific computational patterns of foundational models, particularly transformer architectures, enabling more efficient execution directly within the vehicle by alleviating processing and data transfer bottlenecks.

In practice, compression and specialized hardware are paired with the following runtime tactics to meet strict end-to-end latency requirements:  

\begin{itemize}
\item Run multi-rate asynchronous pipelines keep tracking, free-space, ego-motion fast; offload open-set reasoning and rare-event analysis off the critical path. \cite{liu2022prophet}

\item Use anytime and cascaded inference with early-exit prediction heads and confidence-aware gating to deliver timely outputs and refine when additional time is available. \cite{kaya2019shallow}

\item Maintain streaming and amortized computation by preserving recurrent state, refreshing key frames, and reusing attention Key-Value caches across frames. \cite{liu2023sparsebev, li2024bevformer}

\item Adopt latency-aware architectures with token and voxel pruning, structured sparsity, mixture of experts with capped routing, and localized or windowed attention to bound computational complexity while retaining critical context. \cite{rao2021dynamicvit, huang2024toward}

\end{itemize}
\subsection{\MakeUppercase{Addressing Data Bias and Ensuring Equitable Performance}}

A critical area of focus is the inherent bias present in the vast datasets used to train these models. These datasets often exhibit a systematic bias toward favorable weather conditions and under represent diverse geographic and demographic scenarios, as well as vulnerable road users such as people with disabilities and children. This can lead to models that perform reliably in common situations but fail unexpectedly in edge cases, posing a significant safety risk. Mitigation strategies will require a concerted effort to develop more inclusive and representative datasets through techniques like data augmentation, the generation of high-fidelity synthetic data, and targeted data collection campaigns. Furthermore, research into algorithmic fairness and bias-mitigation techniques during the model training process is essential to ensure that autonomous vehicles can operate safely and equitably in a wide range of real-world environments.

\subsection{\MakeUppercase{The Regulatory Challenge of Non-Deterministic AI}}

The non-deterministic nature of foundation models presents a formidable challenge for regulatory compliance and certification. Traditional automotive safety standards, designed for predictable mechanical systems, are ill-equipped to validate systems that can exhibit emergent behaviors and are not easily interpretable. This "black box" problem is a major hurdle for regulators, who cannot rely on conventional testing protocols to guarantee safety. The current regulatory landscape is a complex patchwork of state-level rules, creating an inconsistent environment for manufacturers. To bridge this gap, future research must focus on developing robust frameworks for the verification and validation of foundation models. A key component of this will be the advancement of Explainable AI (XAI) techniques. By providing clear insights into the model's decision-making process, XAI can help build trust with regulators and the public, facilitate the identification of potential failure modes during audits, provide crucial data for accident investigations, and ultimately pave the way for the safe and responsible deployment of these powerful technologies. This new paradigm of certification will likely combine XAI with large-scale simulation for edge-case testing and continuous real-world data reporting.

\subsection{\MakeUppercase{Mitigating Model Hallucination and Safety Risks}}

Finally, the risk of model hallucination, where a generative model produces outputs that are inconsistent with reality, is a critical safety concern. In the context of autonomous driving, a hallucination could manifest as the perception of a non-existent obstacle or, conversely, the failure to detect a real one, with potentially catastrophic consequences. Mitigating this risk will require a multi-pronged approach. This includes the development of novel architectures that are less prone to hallucination, the creation of comprehensive benchmarks for testing and evaluating model robustness, and the implementation of real-time monitoring systems that can detect and flag potential hallucinations. Ultimately, ensuring the safety of autonomous vehicles powered by foundation models will necessitate a holistic approach that combines rigorous data governance, transparent and interpretable models, and a comprehensive regulatory framework that can adapt to the rapid pace of technological innovation.

\section{\MakeUppercase{Conclusion}}
Foundation models are transforming autonomous driving perception, shifting from specialized algorithms to versatile, large-scale architectures. This fundamental rethinking enables systems to better interpret and anticipate real-world complexities. Pre-trained on vast, diverse datasets, these models address key challenges in scalability, generalization to "long-tail" scenarios, and robust adaptation to changing conditions.

This review has underscored four foundational pillars essential for these systems. Firstly, Generalized Knowledge enables nuanced understanding of novel situations by leveraging Vision Foundation Models (VFMs), Vision Language Models (VLMs), and Large Language Models (LLMs) via methods like distillation and pseudo-labeling. Secondly, Spatial Understanding creates comprehensive 3D environmental representations, capturing geometry and traversable space through volumetric modeling and self-supervised techniques. Thirdly, Multi-Sensor Robustness ensures reliable performance by fusing diverse sensor data (cameras, LiDAR, radar) using cross-modal learning, knowledge distillation, and consistency methods. Lastly, Temporal Understanding allows perception and prediction of scene dynamics and motion for proactive decisions, using 4D prediction and temporal contrastive learning. Mastering these capabilities is crucial for safer, more reliable autonomous systems. Such systems could intuitively handle novel scenarios, navigate adverse conditions without catastrophic perception failure, and better anticipate erratic movements, fulfilling the promise of foundation models.

Despite the immense potential of foundation models, key challenges remain in translating this promise into deployable autonomous driving systems. Beyond high computational demands and the need for seamless sensor integration with robust spatio-temporal alignment and error-resilient fusion, the most critical obstacle lies in the integration of core capabilities—generalized knowledge, spatial reasoning, multi-sensor robustness, and temporal understanding—into a single, efficient, real-time framework. The complexity of unifying these aspects often results in fragmented hybrid systems that trade off full integration for practical viability. While such interim approaches offer short-term utility, achieving truly unified models will require advances not only in model architecture and training strategies, but also in benchmark development that emphasizes hard cases, and system-level design optimized for real-time constraints. Continued progress across these dimensions is essential to fully realize the transformative potential of foundation models in autonomous driving.

\bibliographystyle{IEEEtran}
\bibliography{reference}

\end{document}